\pdfoutput=1
\documentclass[11pt]{article}

\usepackage[final]{acl}

\usepackage{times}
\usepackage{latexsym}

\usepackage[T1]{fontenc}
\usepackage[utf8]{inputenc}

\usepackage{microtype}

\usepackage{inconsolata}

\usepackage{graphicx}
\usepackage{booktabs}
\usepackage{amssymb}
\usepackage{pifont}
\usepackage{amsmath}
\usepackage{multirow}
\usepackage{colortbl}
\usepackage[most]{tcolorbox}
\usepackage{xcolor}
\usepackage{fvextra}
\usepackage{fancyvrb}
\DefineVerbatimEnvironment{CustomVerbatim}{Verbatim}{formatcom=\normalfont, breaklines=true, breaksymbolleft={}, breaksymbolright={}}
\usepackage{ragged2e}
\usepackage[normalem]{ulem}
\useunder{\uline}{\ul}{}

\title{LR$^2$Bench: Evaluating Long-chain Reflective Reasoning Capabilities of Large Language Models via Constraint Satisfaction Problems}

\author{
    Jianghao Chen\textsuperscript{1,2,3}, \ 
    Zhenlin Wei\textsuperscript{1,2},\ 
    Zhenjiang Ren\textsuperscript{1,2}, \
    Ziyong Li\textsuperscript{1,2}, \
    Jiajun Zhang\textsuperscript{1,2,4}\thanks{\ \ Corresponding Author}   \\
    \textsuperscript{1}Institute of Automation, Chinese Academy of Sciences\\
    \textsuperscript{2}School of Artificial Intelligence, University of Chinese Academy of Sciences\\
    \textsuperscript{3}Zhongguancun Academy, Beijing, China \textsuperscript{4}Wuhan AI Research\\
    \texttt{\{chenjianghao2022, weizhenlin2025, renzhenjiang2024, liziyong2023\}@ia.ac.cn}\\
    \texttt{jjzhang@nlpr.ia.ac.cn} \\
}

\begin{document}
\maketitle
\begin{abstract}
Recent progress in Large Reasoning Models (LRMs) has significantly enhanced the reasoning abilities of Large Language Models (LLMs), empowering them to tackle increasingly complex tasks through reflection capabilities, such as making assumptions, backtracking, and self-refinement. However, effectively evaluating such reflection capabilities remains challenging due to the lack of appropriate benchmarks. To bridge this gap, we introduce LR$^2$Bench, a novel benchmark designed to evaluate the \textbf{L}ong-chain \textbf{R}eflective \textbf{R}easoning capabilities of LLMs. LR$^2$Bench comprises 850 samples across six Constraint Satisfaction Problems (CSPs) where reflective reasoning is crucial for deriving solutions that meet all given constraints. Each type of task focuses on distinct constraint patterns, such as knowledge-based, logical, and spatial constraints, providing a comprehensive evaluation of diverse problem-solving scenarios. Our extensive evaluation on both conventional LLMs and LRMs reveals that even the most advanced LRMs, such as DeepSeek-R1 and OpenAI o1-preview, struggle with tasks in LR$^2$Bench, achieving an average Exact Match score of only 20.0\% and 23.6\%, respectively. These findings underscore the significant room for improvement in the reflective reasoning capabilities of current LLMs. \footnote{Our code is available at \url{https://github.com/ZNLP/LR2Bench}}
\end{abstract}

\section{Introduction}

Recent advancements in Large Reasoning Models (LRMs), exemplified by QwQ-32B, DeepSeek-R1, and OpenAI-o1 \citep{qwq-32b-preview, guo2025deepseek, o1}, have demonstrated substantial progress in the reasoning capabilities of Large Language Models (LLMs). These models exhibit more human-like behaviors, such as making assumptions, verification, backtracking, and self-correction, enabling them to address increasingly complex challenges across various domains \citep{zhong2024evaluation, wu2024comparative, wang2024planning, sun2025ktae}.

\begin{figure}[t]
  \centering
  \includegraphics[width=0.9\columnwidth]{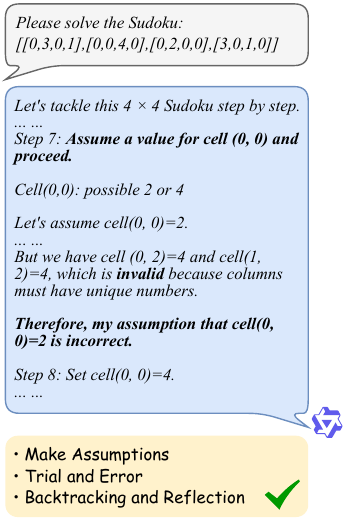}
  \caption{The response demonstrates the model's reflective reasoning process while solving a Sudoku problem.}
  \label{motivation}
  \vspace{-4mm}
\end{figure}

% \begin{table*}[t]
%     \centering
%     {\renewcommand{\arraystretch}{1.0}\resizebox{1.0\textwidth}{!}{\begin{tabular}{llcp{3cm}}
%         \toprule
%         \textbf{Task} & 
%         \textbf{Samples} & 
%         \textbf{Difficulty} & 
%         \textbf{Commonsense Reasoning} \\
%         \midrule
%         Crossword & 5x5, 10x10, 15x15 & 150 & \checkmark \\
%         \bottomrule
%     \end{tabular}}}
%     \caption{\label{Overview}Overview.}
% \end{table*}

\begin{table*}[t]
\centering
{
% \footnotesize
\resizebox{1.0\textwidth}{!}{\begin{tabular}{llccccc}
\toprule
\textbf{Task} &  
\textbf{Difficulty} &
\textbf{Samples} & 
\begin{tabular}[c]{@{}c@{}}\hspace{0.5em} \textbf{Knowledge-based} \hspace{0.5em} \\ \textbf{Constraint} \end{tabular} & 
\begin{tabular}[c]{@{}c@{}}\hspace{0.5em} \textbf{Logical} \hspace{0.5em} \\ \textbf{Constraint} \end{tabular} & 
\begin{tabular}[c]{@{}c@{}}\hspace{0.5em} \textbf{Spatial} \hspace{0.5em} \\ \textbf{Constraint} \end{tabular} & 
\textbf{Reflection} \\
\midrule
Crossword & 5$\times$5, 10$\times$10, 15$\times$15 & 150 & \ding{51} & \ding{55} & \ding{51} & \ding{51} \\
Acrostic & Easy, Hard & 100 & \ding{51} & \ding{55} & \ding{51} & \ding{51} \\
Logic Puzzle & 4$\times$4, 4$\times$5, 4$\times$6, 4$\times$7  & 200 & \ding{55} & \ding{51} & \ding{55} & \ding{51} \\
Cryptogram & Easy, Hard & 100 & \ding{51} & \ding{55} & \ding{55} & \ding{51} \\
Sudoku & 4$\times$4, 9$\times$9 / Easy, Hard & 200 & \ding{55} & \ding{51} & \ding{51} & \ding{51} \\
Drop Quote & Easy, Hard & 100 & \ding{51} & \ding{55} & \ding{51} & \ding{51} \\
\bottomrule
\end{tabular}}}
\caption{Data statistics of LR${}^{2}$Bench.}
\label{tab:overview}
\vspace{-3mm}
\end{table*}

Despite this progress, the reflection capabilities of LLMs remain largely unexplored. Reflection can be defined as the process of engaging in attentive, critical, exploratory, and iterative self-interactions with one's thoughts, actions, and underlying conceptual frameworks \citep{nguyen2014reflection}. Existing research predominantly focuses on evaluating LLMs' abilities to utilize explicit feedback for self-criticism or self-refinement \citep{li2024hindsight, lan2024criticeval, gou2024critic, sun2024critique, lin-etal-2024-criticbench}. However, these approaches overlook a fundamental question: \textbf{\textit{Can LLMs spontaneously engage in the whole reflection process to solve more complex tasks?}} Reflection is not simply about reacting to feedback. It also encompasses exploring possible solutions, assessing and adjusting strategies, and adapting when confronted with contradiction. To better understand how LLMs might engage in these capabilities, we consider Constraint Satisfaction Problems (CSPs). CSPs are defined as a set of variables whose state must satisfy specific constraints, which inherently require iterative exploration in a large search space without predefined solution paths \citep{dechter2003constraint}. As shown in Figure \ref{motivation}, the response of QwQ-32B-Preview \citep{qwq-32b-preview} to Sudoku, a classic CSP, exemplifies a typical reflection process. This task presents a scenario with row, column, and grid constraints, requiring iterative trial-and-error to determine valid values for each cell. The model initially makes an assumption, then identifies a conflict, and finally backtracks to correct its initial guess. This demonstrates an effective reflective reasoning process often absent from standard NLP tasks, such as summarization, translation, and question-answering.

Therefore, to systematically investigate reflective reasoning in LLMs, we propose LR$^2$Bench, a novel benchmark for evaluating the \textbf{L}ong-chain \textbf{R}eflective \textbf{R}easoning capabilities of LLMs. LR$^2$Bench consists of six CSPs: Crossword, Acrostic, Logic Puzzle, Cryptogram, Sudoku, and Drop Quote. Each task necessitates reflection processes and emphasizes specific constraint patterns, such as knowledge-based constraints, logical constraints, and spatial constraints, ensuring a comprehensive assessment of LLMs' reflective reasoning capabilities across diverse problem-solving scenarios. We manually collect and annotate a total of 850 examples spanning multiple difficulty levels across all task types. Through extensive experiments, we observe that even current top-performing LRMs (e.g., DeepSeek-R1 and o1-preview) struggle to complete tasks in LR$^2$Bench, underscoring substantial room for improvement in this critical area. Furthermore, most conventional LLMs exhibit inferior performance to LRMs and achieve near-zero sample-level accuracy in LR$^2$Bench, highlighting the essential role of reflective reasoning skills in tackling the challenges posed by our benchmark.

Our contributions are summarized as follows: 1) We introduce LR$^2$Bench, a novel benchmark with six tasks and diverse difficulty levels, designed to evaluate the long-chain reflective reasoning capabilities of LLMs. 2) We conduct a comprehensive evaluation of reflective reasoning capabilities in both conventional LLMs and LRMs. 3) We present a detailed analysis of the limitations exhibited by current LLMs on LR$^2$Bench, establishing a foundation for future research on the development of advanced reasoning models.

\begin{figure*}[t]
  \includegraphics[width=\linewidth]{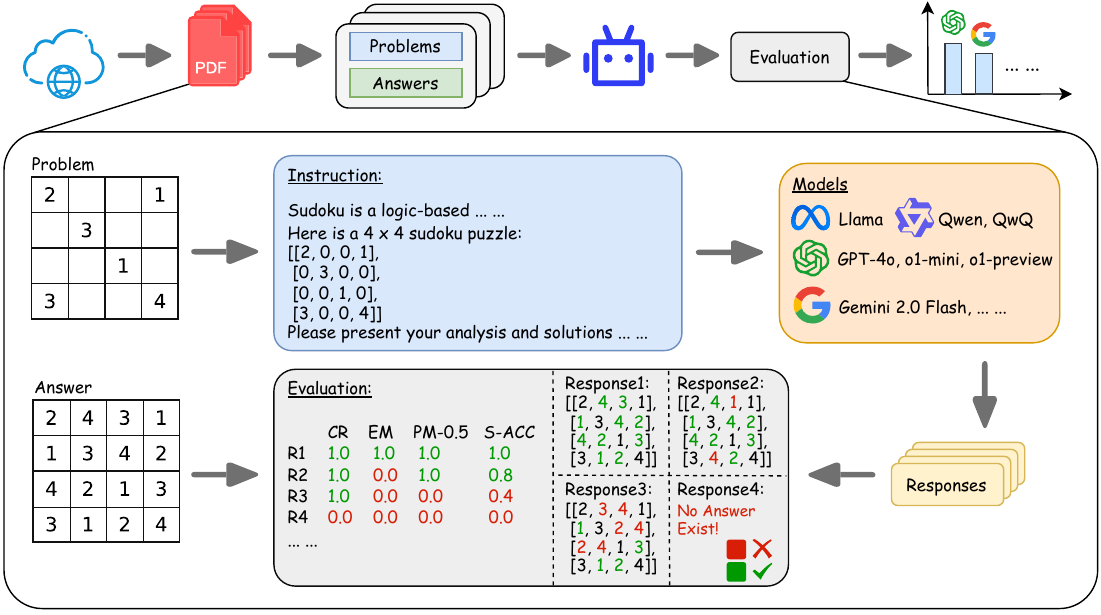}
  \caption{The overall pipeline of LR$^2$Bench, illustrated with the Sudoku task.}
  \label{framework}
  \vspace{-4mm}
\end{figure*}

\section{LR$^2$Bench}
In this section, we introduce the construction process of LR$^2$Bench. Section \ref{sub1} provides an overview of each task. Section \ref{sub2} discusses the diverse scenarios targeted by LR${}^2$Bench. Section \ref{sub3} shows the annotation process and data statistics, and Section \ref{sub4} outlines the evaluation metrics. Figure \ref{framework} illustrates the data collection, annotation, and evaluation pipeline of our LR$^2$Bench.

% \subsection{Preliminary Analysis}

\subsection{Task Overview}
\label{sub1}
% We first introduce the task descriptions and collection methods for all tasks in LR$^2$Bench. For detailed information, please refer to Appendix \ref{app_example}.

We first introduce the task descriptions and collection methods for six CSPs in LR$^2$Bench. To better evaluate the reflection capabilities, we consciously selected CSPs that mainly rely on common knowledge. This minimizes the potential bias of domain-specific expertise and allows us to more rigorously assess the model's ability for iterative exploration and self-correction. For detailed tasks information, please refer to Appendix \ref{app_example}.

\paragraph{Crossword} The Crossword task requires inferring correct words from given clues and filling them into a grid. A key challenge lies in satisfying the constraint of shared letter intersections between horizontal and vertical words. We collected 150 Crossword samples published in 2024 from Los Angeles Times\footnote{\url{https://www.latimes.com}} and Vulture\footnote{\url{https://www.vulture.com}} in three sizes: $5\times5$, $10\times10$, and $15\times15$, with 50 ones for each size.

\paragraph{Acrostic} The Acrostic task involves word clues like Crossword, but its objective is to form a hidden quotation or sentence from the answers to the clues. This requires that the answer words not only satisfy the corresponding clues but also effectively integrate to construct the ultimate hidden message. We collected 50 easy and 50 hard Acrostic samples from Printable Puzzles\footnote{\url{https://www.printable-puzzles.com/printable-acrostic-puzzles.php}} with timestamps ranging from September 2024 to December 2024.

\paragraph{Logic Puzzle} The Logic Puzzle task constitutes a problem that necessitates logical reasoning to deduce relationships between a set of entities based on the given constraints and clues. The objective is to systematically analyze the given information, employing techniques such as hypothesis formation, elimination, and deductive inference, to determine a unique solution that satisfies all given constraints. We collected 50 puzzles for each of the four sizes ($4\times4$, $4\times5$, $4\times6$, and $4\times7$) from Printable Puzzles\footnote{\url{https://www.printable-puzzles.com/printable-logic-puzzles.php}}, with timestamps ranging from September 2024 to December 2024.

\paragraph{Cryptogram} The Cryptogram task involves the decryption of an encrypted quotation or sentence, where each letter of an original text is substituted with another, resulting in an apparently nonsense text. Decryption requires identifying patterns, common letter frequencies, and word structures to deduce the letter-to-letter correspondences, ultimately reconstructing the original content. We collected 50 easy and 50 hard samples from Printable Puzzles\footnote{\url{https://www.printable-puzzles.com/printable-cryptograms.php}} with timestamps ranging from September 2024 to December 2024.

\paragraph{Sudoku} The Sudoku task consists of filling a $n^2 \times n^2$ grid with digits from 1 to $n^2$, subject to the constraint that each row, column, and $n \times n$ subgrid contains all digits from 1 to $n^2$ without repetition. Success in Sudoku relies on logical deduction and careful consideration of the existing digits to determine valid placements for the remaining numbers. From 1sudoku\footnote{\url{https://1sudoku.com}}, we collected 200 Sudoku samples in total: 50 easy and 50 hard samples for both $4\times4$ and $9\times9$ sizes.

\paragraph{Drop Quote} The Drop Quote task comprises a grid of multiple rows and columns, with each column providing a set of candidate letters. The task requires determining the correct row for letters in each column, effectively "dropping" it into target place to reveal the hidden quotation. We created 50 easy samples by manually compiling common quotations, and collected 50 hard samples from Printable Puzzles\footnote{\url{https://www.printable-puzzles.com/printable-drop-quotes.php}}, with timestamps ranging from September 2024 to December 2024.

% \subsection{What Can LR$^2$Bench Tell?}
\subsection{Diverse Scenarios}
\label{sub2}
% We then investigate the specific capabilities required for successfully completing the tasks in LR$^2$Bench. This aims to illustrate the necessary reasoning skills and processing strategies for these tasks, offering diverse perspectives for further evaluation of LLMs.

Each type of task within LR$^2$Bench focuses on different constraint patterns, providing a comprehensive framework to evaluate models’ reflective reasoning capabilities across diverse scenarios. We further explore the varying capabilities required for completing the tasks in LR$^2$Bench.

\paragraph{Reflection} Reflection is the most fundamental capability for tackling the complex tasks in LR$^2$Bench. Unlike simple problems with short-cut solutions, these tasks are inherently iterative, demanding the exploration of multiple possibilities, identification of dead ends, and adaptive revision of initial hypotheses. Such reflective reasoning capabilities enable a thorough analysis and refinement of strategies, ultimately leading to more robust and effective solutions.

\paragraph{Long-chain Generation} LR$^2$Bench incorporates tasks that necessitate long-chain generation, a crucial capability for LLMs to tackle complex reasoning problems. Unlike tasks with simple and isolated answers, these tasks require LLMs to generate a long chain of steps or decisions that build upon each other toward a final solution.

\paragraph{Knowledge-based Reasoning} Both Crossword and Acrostic tasks demand broad world knowledge and commonsense reasoning abilities since the clue answers often hinge on cultural references, idiomatic expressions, and diverse factual domains. Additionally, the Cryptogram and Drop Quote tasks require knowledge of typical phrase structures and common linguistic patterns to decode messages or reconstruct quotations effectively.

\paragraph{Logical Reasoning} The logical reasoning ability is essential across various tasks, especially evident in the Logic Puzzles and Sudoku. These tasks involve information integration and systematic application of deductive reasoning to solve problems constrained by specific rules.

\paragraph{Spatial Reasoning} While not the primary focus for all tasks, spatial reasoning also emerges as a critical capability within LR$^2$Bench, particularly in tasks with grids. For Crossword, considering letter intersections across horizontal and vertical placements is crucial. Similarly, in Sudoku, effective digit placement requires reasoning about row, column, and subgrid constraints, all of which involve spatial relationships within the grid.

\subsection{Data Annotation and Statistics}
\label{sub3}
For PDF data collected from websites, we manually convert key elements in task samples into a structured text format suitable for LLMs' inputs. Then, we manually craft task-specific instructions, including problem definitions and rules, to provide LLMs with the necessary guidance to process each task effectively. To further control the output format of LLMs, we manually construct two simple shots for each type of task to facilitate subsequent answer extraction and evaluation. Table \ref{tab:overview} presents all tasks in LR$^2$Bench along with their attributes. Appendix \ref{app_example} shows the text-format task examples. Appendix \ref{app_prompt} shows the detailed instructions and few-shot examples for each type of task. Appendix \ref{app_annotation} details the annotation specification and task allocation.

\subsection{Evaluation Metrics}
\label{sub4}
Since all tasks in LR$^2$Bench consist of multiple subtasks (e.g., words inference in Crossword, cells completion in Sudoku), we employ fine-grained, subtask-level evaluation metrics. Appendix \ref{app_subtask} illustrates the detailed subtask definition. Given a problem with $N$ subtasks, let $G = \{ g_1, \ldots, g_N \}$ and $P = \{ p_1, \ldots, p_N \}$ denote the ground truth and LLM-generated answers for each subtask, respectively. We define the following evaluation metrics (refer to Figure \ref{framework} for a concrete illustration of these metrics applied to Sudoku tasks):

\paragraph{Completion Ratio} The Completion Ratio (CR) metric measures the proportion of subtasks within a given problem that LLMs successfully complete, regardless of the correctness of the answers. 
CR is calculated as follows:
\begin{equation}\label{cr}
\text{CR} = \frac{1}{N} \sum_{i=1}^{N} \mathbb{I}(p_i \neq \varnothing)
\end{equation}
where $\mathbb{I}(p_i \neq \varnothing)$ equals 1 if subtask $i$ is completed (i.e., $p_i$ is not empty) and 0 otherwise. In our experiments, we observe that some models struggle to produce a complete reasoning chain and fail to reach the final answer for each subtask. Therefore, we propose CR to measure the long reasoning chain generation capability of LLMs.

\paragraph{Subtask Accuracy} For fine-grained evaluation, we propose Subtask Accuracy (S-Acc), which calculates the proportion of correctly solved subtasks compared to the ground truth:
\begin{equation}\label{s-acc}
\text{S-Acc} = \frac{1}{N} \sum_{i=1}^{N} \mathbb{I}(p_i = g_i)
\end{equation}

\paragraph{Exact Match} The Exact Match (EM) metric employs a strict correctness criterion: for each subtask, the generated answer by LLMs must exactly match the ground truth. EM is calculated as follows:
\begin{equation}\label{em}
\text{EM} = \mathbb{I}\left( \frac{1}{N} \sum_{i=1}^{N} \mathbb{I}(p_i = g_i) = 1.0 \right)
\end{equation}
This metric is crucial for evaluating the absolute accuracy on individual samples. For example, in a Sudoku task, EM would require that the completed grid be identical to the ground truth, with no errors in any row, column, or subgrid.

\paragraph{Partial Match} Considering that our tasks often involve multiple interconnected subtasks, making a completely correct solution (as measured by EM) exceptionally challenging, we introduce a Partial Match metric (PM-0.5):
\begin{equation}\label{pm05}
\text{PM-0.5} = \mathbb{I}\left( \frac{1}{N} \sum_{i=1}^{N} \mathbb{I}(p_i = g_i) \geq 0.5 \right)
\end{equation}
By analyzing PM-0.5 scores, we can differentiate between models that consistently fail to make progress on the tasks and those that can correctly solve a significant fraction of the sub-problems. A higher PM-0.5 score (compared to a low EM score) suggests that the model possesses some level of the required reasoning skills but struggles with maintaining consistency or navigating the entire solution space without errors.

% Reasoning Depth 对具体推理路径的分析指标？
% Time to Solution output tokens?

\section{Experiments}

\subsection{Experimental Setup}
% todo
\paragraph{Evaluated Models} We evaluate 11 open-source models and 5 closed-source models. Specifically, we include leading LRMs: QwQ-32B-Preview \citep{qwq-32b-preview}, DeepSeek-R1 \citep{guo2025deepseek}, Gemini-2.0-flash-thinking \citep{deepmind_gemini_flash_thinking}, OpenAI o1-mini and o1-preview \citep{o1}. Please refer to Appendix \ref{app_models} for detailed information of all selected models.

\paragraph{Implementation Details} We utilize the default prompt templates for all LLMs. Appendix \ref{app_example} and \ref{app_prompt} show detailed problem formats, instructions, and few-shot examples for each type of task. Appendix \ref{app_shot} discuss the influence of different types of few-shot samples. To facilitate evaluation, we wrap answers within specific tags (e.g., \texttt{<Answer>} and \texttt{</Answer>}) in the few-shot examples. This enables precise answer extraction from the model responses. Then the extracted answer (a string) is parsed into a structured format for further metrics calculation. We use the vLLM inference framework \citep{kwon2023efficient} and employ greedy sampling with temperature = 0 to minimize randomness, except for o1-mini and o1-preview which have an inherent temperature of 1. The maximum sequence length is set to the default maximum value for each model.

\subsection{Main Results}

Table \ref{tab:main_exp_sum} presents the average performance across six tasks on LR$^2$Bench, with individual task results detailed in Table \ref{tab:main_exp}.

\begin{table}[!h]
    \centering
    \resizebox{1.0\columnwidth}{!}{
    \begin{tabular}{l|ccccc}
        \toprule
        \textbf{Model} & \textbf{CR} & \textbf{S-Acc} & \textbf{EM} & \textbf{PM-0.5} & \textbf{\# Tokens} \\
        \midrule
        \rowcolor{gray!15} \multicolumn{6}{l}{\textit{\textbf{Open-source LLMs}}}\\
        \midrule 
        Llama-3.1-8B-Instruct & 42.6 & 9.9 & 0.0 & 3.8 & 2,478 \\
        Llama-3.1-70B-Instruct & 71.8 & 27.4 & 0.5 & 21.9 & 2,090 \\
        Llama-3.3-70B-Instruct & 92.4 & 33.1 & 1.3 & 25.8 & 1,842 \\
        Mistral-7B-Instruct-v0.3 & 85.8 & 12.1 & 0.0 & 2.3 & 2,736 \\
        Mistral-Small-Instruct-2409 & 91.0 & 23.1 & 0.2 & 13.3 & 2,273 \\
        Mistral-Large-Instruct-2411 & 96.1 & \underline{36.4} & 2.5 & \underline{30.0} & 2,313 \\
        Qwen2.5-7B-Instruct & 85.1 & 17.7 & 0.3 & 5.1 & 2,086 \\
        Qwen2.5-32B-Instruct & \underline{96.2} & 29.9 & 0.6 & 14.8 & 1,924 \\
        Qwen2.5-72B-Instruct & 95.0 & 33.9 & 0.9 & 20.8 & 2,359 \\
        QwQ-32B-Preview & 65.0 & 26.6 & \underline{8.5} & 19.3 & 6,709 \\
        DeepSeek-R1 & \textbf{100.0} & \textbf{58.4} & \textbf{20.0} & \textbf{62.0} & 9,856 \\
        \midrule
        \rowcolor{blue!15} \multicolumn{6}{l}{\textit{\textbf{Closed-source LLMs}}}\\
        \midrule
        Gemini-2.0-flash & 81.1 & 37.0 & 2.4 & 34.5 & 2,637 \\
        Gemini-2.0-flash-thinking & 88.2 & 39.4 & 4.3 & 35.0 & 3,725 \\
        GPT-4o & \textbf{99.8} & \underline{43.7} & 3.2 & \underline{41.7} & 1,486 \\
        o1-mini & \underline{97.7} & 41.3 & \underline{9.1} & 32.8 & 9,576 \\
        o1-preview & 96.3 & \textbf{58.7} & \textbf{23.6} & \textbf{61.7} & 11,436 \\ 
        \bottomrule
    \end{tabular}}    
    \caption{Average performance (\%) across six tasks on LR${}^{2}$Bench. The best and second-best results are highlighted in \textbf{bold} and \underline{underlined}, respectively. "\# Tokens" denotes the average number of generated tokens.}
    \label{tab:main_exp_sum}
    % \vspace{-2mm}
\end{table}
\begin{table*}[!t]
    \centering
    \renewcommand{\arraystretch}{1.0}\resizebox{1.0\textwidth}{!}{
    \setlength{\tabcolsep}{3pt}
    \begin{tabular}{l|ccccc|ccccc|ccccc}
        \toprule
        \multirow{2}{*}{\textbf{Model}} &  \multicolumn{5}{c|}{\textbf{Crossword}} & \multicolumn{5}{c|}{\textbf{Acrostic}}  & \multicolumn{5}{c}{\textbf{Logic Puzzle}} \\
        \cmidrule(l){2-16}
        & CR & S-Acc & EM & PM-0.5 & \# Tokens & CR & S-Acc & EM & PM-0.5 & \# Tokens & CR & S-Acc & EM & PM-0.5 & \# Tokens \\
        \midrule
        \rowcolor{gray!15} \multicolumn{16}{l}{\textit{\textbf{Open-source LLMs}}}\\
        \midrule 
        % \rowcolor{blue!15} GPT-4o-2024-08-06 & 54.82 & 898 & 29.71 & 457 & \textbf{42.88} & 1436 & \textbf{67.49} & 1581 & 47.87 & 1121 & \textbf{48.55} & 1098 \\
        Llama-3.1-8B-Instruct & 61.3 & 23.3 & 0.0 & 14.0 & 2,888 & 43.0 & 5.6 & 0.0 & 0.0 & 3,712 & 57.0 & 16.0 & 0.0 & 8.0 & 1,293 \\
        Llama-3.1-70B-Instruct & 77.3 & 46.8 & 0.0 & 62.0 & 3,072 & 84.0 & 35.8 & 0.0 & 21.0 & 3,565 & 56.0 & 22.8 & 2.0 & 18.0 & 1,165 \\
        Llama-3.3-70B-Instruct & 85.3 & 47.6 & 0.0 & 65.3 & 2,613 & 97.0 & \underline{40.8} & 0.0 & \underline{28.0} & 3,584 & 80.5 & 32.2 & 1.0 & 25.0 & 1,738 \\
        Mistral-7B-Instruct-v0.3 & 94.0 & 23.0 & 0.0 & 6.7 & 3,655 & 75.0 & 7.9 & 0.0 & 0.0 & 4,600 & 97.0 & 19.1 & 0.0 & 4.5 & 1,618 \\
        Mistral-Small-Instruct-2409 & 98.7 & 48.3 & 0.0 & 54.0 & 3,135 & 67.0 & 5.5 & 0.0 & 0.0 & 4,171 & \underline{99.5} & 30.7 & 0.5 & 12.5 & 1,514 \\
        Mistral-Large-Instruct-2411 & \underline{99.3} & \underline{62.8} & \underline{2.0} & \underline{86.0} & 3,237 & \underline{98.0} & 39.4 & 0.0 & 20.0 & 4,279 & \textbf{100.0} & 38.3 & 3.0 & 30.5 & 1,637 \\
        Qwen2.5-7B-Instruct & 98.7 & 21.1 & 0.0 & 3.3 & 2,441 & 42.0 & 3.6 & 0.0 & 0.0 & 4,159 & 96.5 & 25.8 & 0.0 & 8.5 & 1,396 \\
        Qwen2.5-32B-Instruct & \textbf{100.0} & 34.6 & 0.0 & 20.0 & 2,560 & \textbf{100.0} & 31.8 & 0.0 & 2.0 & 4,073 & 93.0 & 32.2 & 0.0 & 22.5 & 1,208 \\
        Qwen2.5-72B-Instruct & \textbf{100.0} & 44.1 & 0.0 & 36.7 & 2,735 & \textbf{100.0} & 39.3 & 0.0 & 18.0 & 4,111 & 93.5 & 34.0 & 0.0 & 23.0 & 1,810 \\
        QwQ-32B-Preview & 80.0 & 30.2 & 0.0 & 18.0 & 4,817 & 97.0 & 31.6 & 0.0 & 6.0 & 4,964 & 78.5 & \underline{46.3} & \underline{19.5} & \underline{48.0} & 9,524 \\
        DeepSeek-R1 & \textbf{100.0} & \textbf{75.4} & \textbf{16.7} & \textbf{94.0} & 9,810 & \textbf{100.0} & \textbf{62.2} & 0.0 & \textbf{83.0} & 10,077 & \textbf{100.0} & \textbf{69.4} & \textbf{42.5} & \textbf{68.0} & 9,205\\
        \midrule
        \rowcolor{blue!15} \multicolumn{16}{l}{\textit{\textbf{Closed-source LLMs}}}\\
        \midrule
        Gemini-2.0-flash & \underline{98.7} & 61.6 & 0.0 & 83.3 & 2,555 & \underline{98.0} & 48.0 & 0.0 & 48.0 & 4,020 & 58.0 & 24.2 & 2.0 & 20.0 & 2,104 \\
        Gemini-2.0-flash-thinking & 94.7 & 57.7 & \underline{1.3} & 79.3 & 2,648 & 92.0 & 40.7 & 0.0 & 27.0 & 4,257 & \underline{99.0} & 45.9 & 8.0 & 37.5 & 4,038 \\
        GPT-4o & \textbf{100.0} & \underline{66.0} & \underline{1.3} & \underline{86.7} & 1,726 & \textbf{100.0} & \underline{56.0} & 0.0 & \underline{67.0} & 3,229 & \textbf{100.0} & 39.3 & 3.5 & 29.5 & 953 \\
        o1-mini & 95.3 & 45.5 & \underline{1.3} & 54.0 & 7,840 & 97.0 & 34.7 & 0.0 & 12.0 & 10,952 & \underline{99.0} & \underline{57.2} & \underline{23.5} & \underline{53.5} & 10,242 \\
        o1-preview & 98.0 & \textbf{77.7} & \textbf{24.7} & \textbf{89.3} & 10,098 & \textbf{100.0} & \textbf{67.2} & 0.0 & \textbf{90.0} & 14,847 & \underline{99.0} & \textbf{68.8} & \textbf{41.0} & \textbf{68.5} & 9,449 \\ 
        \bottomrule
    \end{tabular}}
\end{table*}
\begin{table*}[!t]
    \centering
    \renewcommand{\arraystretch}{1.0}\resizebox{1.0\textwidth}{!}{
    \setlength{\tabcolsep}{3pt}
    \begin{tabular}{l|ccccc|ccccc|ccccc}
        \toprule
        \multirow{2}{*}{\textbf{Model}} &  \multicolumn{5}{c|}{\textbf{Cryptogram}} & \multicolumn{5}{c|}{\textbf{Sudoku}}  & \multicolumn{5}{c}{\textbf{Drop Quote}} \\
        \cmidrule(l){2-16}
        & CR & S-Acc & EM & PM-0.5 & \# Tokens & CR & S-Acc & EM & PM-0.5 & \# Tokens & CR & S-Acc & EM & PM-0.5 & \# Tokens \\
        \midrule
        \rowcolor{gray!15} \multicolumn{16}{l}{\textit{\textbf{Open-source LLMs}}}\\
        \midrule 
        % \rowcolor{blue!15} GPT-4o-2024-08-06 & 54.82 & 898 & 29.71 & 457 & \textbf{42.88} & 1436 & \textbf{67.49} & 1581 & 47.87 & 1121 & \textbf{48.55} & 1098 \\
        Llama-3.1-8B-Instruct & 43.0 & 2.3 & 0.0 & 0.0 & 2,068 & 7.5 & 1.2 & 0.0 & 0.0 & 2,782 & 44.0 & 11.2 & 0.0 & 1.0 & 2,123 \\
        Llama-3.1-70B-Instruct & 62.0 & 6.9 & 0.0 & \underline{1.0} & 1,298 & 69.5 & 24.2 & 1.0 & 17.5 & 1,940 & 82.0 & 27.7 & 0.0 & 12.0 & 1,498 \\
        Llama-3.3-70B-Instruct & \underline{99.0} & \underline{14.3} & 0.0 & \underline{1.0} & 1,137 & 93.5 & 34.8 & 7.0 & 22.5 & 1,062 & \underline{99.0} & 29.0 & 0.0 & 13.0 & 918 \\
        Mistral-7B-Instruct-v0.3 & \underline{99.0} & 4.3 & 0.0 & 0.0 & 1,096 & 84.0 & 11.9 & 0.0 & 1.5 & 3,108 & 66.0 & 6.6 & 0.0 & 1.0 & 2,337 \\
        Mistral-Small-Instruct-2409 & 95.0 & 7.0 & 0.0 & 0.0 & 1,233 & 89.0 & 20.5 & 0.5 & 7.5 & 1,968 & 97.0 & 26.9 & 0.0 & 6.0 & 1,615 \\
        Mistral-Large-Instruct-2411 & 96.0 & 13.7 & 0.0 & \underline{1.0} & 1,204 & 85.5 & 39.5 & 10.0 & 33.5 & 1,955 & 98.0 & 24.7 & 0.0 & 9.0 & 1,566 \\
        Qwen2.5-7B-Instruct & 81.0 & 3.5 & 0.0 & 0.0 & 1,181 & 94.5 & 30.2 & 1.5 & 15.0 & 1,486 & 98.0 & 21.9 & 0.0 & 4.0 & 1,852 \\
        Qwen2.5-32B-Instruct & 89.0 & 9.8 & 0.0 & 0.0 & 1,303 & \textbf{100.0} & 42.8 & 3.5 & 30.5 & 1,202 & 95.0 & 28.4 & 0.0 & \underline{14.0} & 1,197 \\
        Qwen2.5-72B-Instruct & 85.0 & 11.8 & 0.0 & 0.0 & 1,727 & \underline{97.5} & \underline{43.0} & 5.5 & 34.0 & 2,013 & 94.0 & \underline{30.9} & 0.0 & 13.0 & 1,757 \\
        QwQ-32B-Preview & 47.0 & 3.6 & 0.0 & 0.0 & 6,492 & 54.5 & 40.1 & \underline{31.5} & \underline{35.5} & 8,381 & 33.0 & 7.5 & 0.0 & 8.0 & 6,078 \\
        DeepSeek-R1 & \textbf{100.0} & \textbf{26.0} & \textbf{4.0} & \textbf{21.0} & 10,344 & \textbf{100.0} & \textbf{70.3} & \textbf{50.0} & \textbf{64.0} & 8,277 & \textbf{100.0} & \textbf{47.3} & \textbf{7.0} & \textbf{42.0} & 11,422 \\
        \midrule
        \rowcolor{blue!15} \multicolumn{16}{l}{\textit{\textbf{Closed-source LLMs}}}\\
        \midrule
        Gemini-2.0-flash & 47.0 & 8.5 & 0.0 & 1.0 & 1,585 & 93.0 & 45.3 & 12.5 & 37.5 & 2,843 & 92.0 & 34.3 & 0.0 & 17.0 & 2,717 \\
        Gemini-2.0-flash-thinking & 68.0 & 11.2 & 0.0 & 2.0 & 4,167 & 79.5 & 46.5 & 16.5 & 41.0 & 3,853 & 96.0 & \underline{34.4} & 0.0 & \underline{23.0} & 3,386 \\
        GPT-4o & \textbf{100.0} & 20.7 & 0.0 & 5.0 & 740 & \textbf{100.0} & 52.2 & 14.5 & \underline{48.0} & 1,104 & \textbf{99.0} & 31.1 & 0.0 & 14.0 & 1,165 \\
        o1-mini & \textbf{100.0} & \underline{22.7} & \underline{1.0} & \underline{13.0} & 11,208 & \underline{99.0} & \underline{53.4} & \underline{27.0} & 43.0 & 3,961 & 96.0 & 34.3 & \underline{2.0} & 21.0 & 13,255 \\
        o1-preview & \underline{92.0} & \textbf{34.8} & \textbf{13.0} & \textbf{29.0} & 12,567 & 91.5 & \textbf{65.1} & \textbf{50.0} & \textbf{55.5} & 8,062 & \underline{97.0} & \textbf{38.8} & \textbf{13.0} & \textbf{38.0} & 13,595 \\ 
        \bottomrule
    \end{tabular}}    
    \caption{Performance (\%) of LLMs on six tasks. The best and second-best results are highlighted in \textbf{bold} and \underline{underlined}, respectively. "\# Tokens" denotes the average number of generated tokens.}
    \label{tab:main_exp}
    % \vspace{-2mm}
\end{table*}

As shown in Table \ref{tab:main_exp_sum}, a significant gap persists between EM and other evaluation metrics across all models. This observation indicates that while LLMs can address specific aspects of a given task, achieving a complete and accurate solution remains a significant challenge. Moreover, closed-source models generally exhibit superior performance across key metrics (S-Acc, EM, and PM-0.5) to open-source models, with OpenAI o1-preview achieving the best EM score of 23.6\%. DeepSeek-R1 demonstrates the most advanced reasoning capabilities among open-source models and even achieves a comparable EM score of 20.0\% to OpenAI o1-preview. Notably, there also exists a great performance gap between conventional LLMs and LRMs. Although GPT-4o shows the second-highest S-Acc, it lags significantly behind the top-performing model OpenAI o1-preview in EM. This highlights the effectiveness of the "slow thinking" approaches which trade better performance with more inference tokens.

% For open-source models, larger-scale models shows better reasoning performance. Mistral-Large-Instruct-2411, a dense LLM of 123B parameters, achieves the highest S-Acc and PM-0.5 scores. Consistent with the trend observed in closed-source models, o1-like model QwQ-32B-Preview attains the highest EM score. However, its low CR (Completion Ratio) results in comparatively lower scores on other metrics. Overall, the results indicate that even the top-performing models struggle to perfectly solve these tasks, highlighting the significant room for improvement in such complex reflective reasoning capabilities.

\section{Analysis}
\begin{figure}[!h]
  \centering
  % \vspace*{1mm}
  \includegraphics[width=0.8\columnwidth]{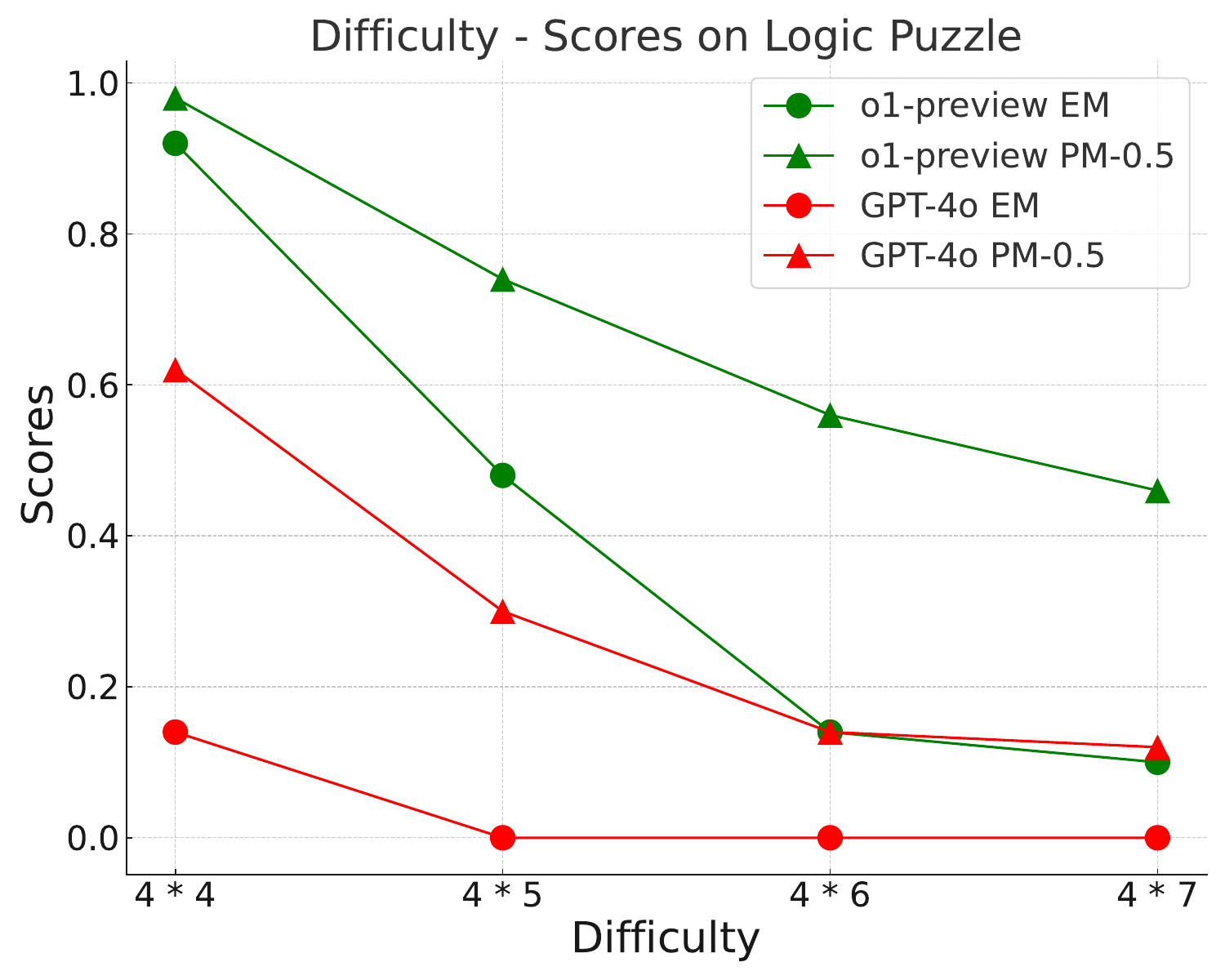}
  \caption{\label{difficulty}Performance trend of GPT-4o and o1-preview under varying difficulty levels.}
\end{figure}
\subsection{Task Difficulty}

We present the detailed model performance across varying difficulty for all six tasks in Appendix \ref{app_difficulty}. Here, we focus on the Logic Puzzle, illustrating the impact of increasing problem complexity. Figure \ref{difficulty} shows the performance of GPT-4o and o1-preview under different difficulty levels. The substantial performance gap between the two models demonstrates o1-preview's superior logical deduction and reflective reasoning abilities. As the solution space expands from $4\times4$ to $4\times7$, while models can still solve part of the problem, they fail to find the complete solution. This highlights the challenge of exploring exponentially increasing solution spaces for complex reasoning tasks.

\subsection{Long Reasoning Chain Generation}

\begin{figure}[h!]
  \centering
  \includegraphics[width=1.0\columnwidth]{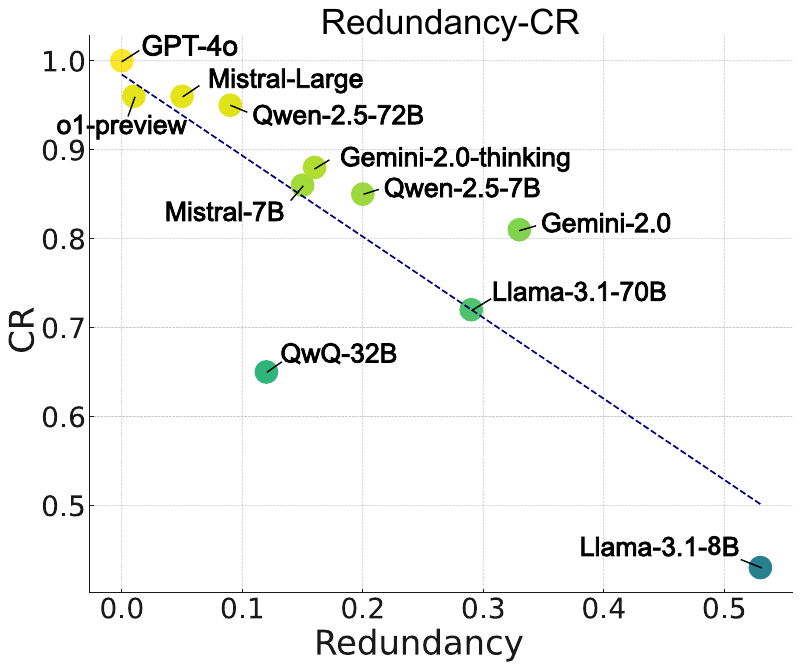}
  \caption{The relationship between redundancy and CR across different models.}
  \label{redundant}
  \vspace{-2mm}
\end{figure}

Tasks in LR$^2$Bench often involve multiple subtasks (e.g., solving all clues for Crossword, inferring the digits for each cell for Sudoku). Such inherent complexity requires LLMs to continuously explore quite large solution spaces through reflective reasoning processes, thus presenting a significant challenge for LLMs in generating long reasoning chains. As shown in Table \ref{tab:main_exp_sum}, many models failed to completely generate the entire reasoning process, resulting in low CR scores. Through analysis of incomplete model responses, we find that a key obstacle to this phenomenon is the tendency of LLMs to generate repetitive content, ultimately reaching the maximum sequence length. This redundancy primarily occurs when LLMs encounter contradictions (see Section \ref{qualitative} for a detailed analysis). Such redundancy wastes valuable context window size, preventing the model from exploring the full solution space and completing further reasoning processes. To quantify this, we calculate the average 10-gram redundancy ratio of the models' responses across all tasks in LR$^2$Bench, excluding Sudoku due to its inherently repetitive cell-by-cell reasoning strategy. Figure \ref{redundant} reveals a strong negative correlation between redundancy and CR, suggesting that redundant generation is a key factor limiting the long-chain reasoning capability of LLMs. Notably, QwQ-32B-Preview shows lower redundancy but still fails in completion due to its endless trial-and-error without reaching a meaningful conclusion.

\subsection{Conventional LLMs vs. LRMs}
% different tasks, difficulty
As shown in Table \ref{tab:main_exp}, for Crossword and Acrostic tasks, we observe that conventional models can achieve high S-Acc and PM-0.5, but nearly zero EM. This suggests that LLMs with strong commonsense reasoning abilities can correctly infer parts of words from the provided clues. However, some clues may have multiple valid answers, requiring LLMs to determine the final answers based on the constraint of shared letters with other words. Therefore, the lack of reflective reasoning prevents these models from achieving perfect solutions. Besides, GPT-4o performs better than o1-mini on these tasks, suggesting that only models possessing both capabilities can achieve high EM scores. 

For tasks that heavily rely on reflection capabilities, such as Logic Puzzle and Sudoku, the performance gap between LRMs and conventional LLMs is the most obvious. Notably, o1-preview outperforms GPT-4o by 37.5\% and 35.5\% and QwQ-32B-Preview outperforms Qwen-2.5-72B-Instruct by 19.5\% and 26.0\% on EM for these two tasks, respectively. These substantial gains highlight the significant advantage of LRMs in tasks that require verifying multiple constraints and employing backtracking mechanisms. In contrast, conventional models often exhibit limitations with their single-pass reasoning paradigm. This approach proves inadequate for scenarios requiring exploration of alternative solution paths or correction of initial assumptions.

Cryptogram and Drop Quote pose the most significant challenges for all evaluated models. We attribute this difficulty to the expansive search space and limited helpful clues inherent in these problems. In the Cryptogram, each letter has theoretically 26 possible substitutions. Similarly, in Drop Quote, each letter within a given column can be dropped to any row. This contrasts with Logic Puzzles and Sudoku, which offer strong constraints that effectively reduce the search space. Even the most advanced model o1-preview struggles to complete these tasks with only a 13\% EM score.

\subsection{Qualitative Analysis}
\label{qualitative}
We analyze several typical behaviors of LLMs leading to the failure in completing our benchmark and provide detailed cases in Appendix \ref{app_case}.

\paragraph{Lack of Reflection Mechanism} This deficiency mainly occurs in conventional models. Taking the Logic Puzzle for example, our analysis of GPT-4o's responses in Appendix \ref{lack} reveals that although the model can effectively break down individual clues and generate initial deductions, it fails to perform the iterative cross-checking to ensure consistency with all established constraints. These shortages indicate that conventional models lack the reflective reasoning capabilities necessary to solve complex constraint satisfaction problems.

\paragraph{Stuck in Contradictions} We observe that LLMs often struggle with complex reasoning tasks when confronted with contradictions. As shown in Appendix \ref{stuck}, this phenomenon mainly manifests as looping within similar sentences, repeatedly stating conflicting information without making progress toward a coherent resolution. Instead of backtracking and revising previous assumptions upon encountering a contradiction, they tend to focus only on the conflicting points. This narrow focus prevents them from exploring alternative solution paths.

\paragraph{Giving-up Moment} As illustrated in Appendix \ref{giveup}, the occurrence of a "Giving-up Moment" is observed when reasoning-specific LLMs struggle with complex problems due to the time (inference tokens) constraint. We focus on one of the worst-performing tasks Cryptogram to quantify this giving-up ratio in QwQ-32B and DeepSeek-R1 responses, respectively. We manually review 50 responses from each model on Cryptogram. For QwQ-32B, 15 out of 50 responses are incomplete due to the maximum context window size of 32k. 26 out of 50 responses show model's giving-up and only provide part of the answers. Similarly, for DeepSeek-R1, 39 out of 50 responses compromise to give the most possible answers given the time constraints. This highlights the challenges posed by problems requiring extensive reflective reasoning and suggests a potential bottleneck in the current capabilities of even the leading reasoning LLMs.

\section{Related Work}
% \subsection{Reasoning with LLMs}
% \subsection{Reflective Reasoning Evaluation}
\subsection{Reflection Capabilities of LLMs}
Recent studies have made significant progress in the long-chain reasoning ability of LLMs, such as long-context understanding \citep{gao2024train, fu2024data, chen2025ladm} and long-cot generation \citep{guo2025deepseek, yeotong2025longcot, zeng2025simplerlzooinvestigatingtamingzero}. These impressive reasoning capabilities have naturally led to increased interest in reflection, a more sophisticated and human-like form of reasoning. Previous studies primarily focus on evaluating LLMs' abilities to rectify their response based on explicit feedback, including self-improvement with critique prompts \citep{lan2024criticeval, li2024reflection, lin-etal-2024-criticbench, li2024hindsight, madaan2024self}, leveraging external tools such as code interpreters or search engines \citep{gou2024critic, chen2024teaching, shinn2024reflexion}, and engaging in multi-LLM interaction through debating \citep{liang-etal-2024-encouraging, huang2024large}. However, these works mainly evaluate LLMs' behaviors in response to feedback. They fail to assess LLMs' capabilities to spontaneously engage in the complete reflection process for complex reasoning tasks. Our proposed LR$^2$Bench provides scenarios necessitating capabilities, such as making assumptions, verification, backtracking, and self-refinement, thus filling a critical gap in evaluating LLMs' intrinsic reflective reasoning abilities.

% As LLMs demonstrate improved reasoning capabilities, current research focuses on more human-like abilities like reflection.

% CRITICEVAL evaluates LLMs' critique ability across four dimensions. CRITICBENCH analyzes the relationship between the generation, critique, and correction capability of LLMs. Reflection-Bench evaluates the intelligence of LLMs through core cognitive functions for reflection. Other works utilize explicit feedback mechanisms, such as critique prompts, model feedback, and external tools, to evaluate the reflection capabilities of LLMs. However, these studies fail to evaluate LLMs' capabilities to spontaneously engage in the whole reflection process for complex reasoning tasks. They either  分开评估批判 修正等或依赖外部，缺乏整体 

% \subsection{Complex Reasoning Tasks for LLMs}
\subsection{Puzzle-solving for LLMs}
% Evaluating LLMs across diverse scenarios is crucial for understanding their complex reasoning capabilities. 
Puzzle-solving \citep{giadikiaroglou2024puzzle} offers valuable insight for evaluating the complex reasoning capabilities of LLMs across diverse scenarios. \citet{ishay2023leveraging} explore Sudoku solving strategies with answer set programming. \citet{ding2023everything} leverages reinforcement learning and Monte Carlo Tree Search to solve problems like Game of 24, 8-Puzzle, and Pocket Cube. \citet{yao2024tree} introduces "Tree of Thought" to enable self-evaluating and backtracking for Game of 24 and Crosswords. \citet{mittal2024puzzlebench} combines LLMs with symbolic solvers and program interpreters to complete first-order combinatorial reasoning problems. \citet{tyagi2024step} focuses on grid puzzles to evaluate the generated reasoning chains of LLMs. Moreover, existing studies have also investigate Board game \citep{kazemi2024boardgameqa}, Chess \citep{feng2024chessgpt} and social games \citep{light2023avalonbench, wang2023avalon, xu2023exploring}. However, these studies primarily leverage external tools or specialized algorithms to develop task-specific solutions within limited puzzle domains. In contrast, LR$^2$Bench provides diverse tasks and difficulty levels and focuses on evaluating the intrinsic reflective reasoning capabilities of LLMs.
% Unlike previous studies that focus on task-specific solutions of limited puzzle domains, LR$^2$Bench provides diverse tasks and difficulty levels and targets evaluating the intrinsic reflective reasoning capabilities of LLMs.
% However, these studies typically focus on task-specific solutions rather than comprehensive evaluation over diverse LLMs. With the advancements of o1-like models, LLMs have demonstrated strong capabilities to independently complete these complex reasoning tasks. Therefore, we propose LR$^2$Bench to provide a comprehensive evaluation and comparison of LLMs across a range of tasks.
% 

\section{Conclusion}
This paper introduces LR$^2$Bench, a novel benchmark to comprehensively evaluate the reflection capabilities of LLMs in long-chain reasoning. LR$^2$Bench comprises six tasks with varying difficulty levels, providing a thorough analysis across diverse scenarios. The experimental results show that LRMs outperform conventional LLMs, demonstrating their superior performance on reflective reasoning. Our findings also highlight the limitation of current reasoning LLMs and reveal that even the most advanced reasoning models fall short of achieving satisfactory performance, suggesting significant room for enhancement in reflective reasoning capabilities.

\section*{Limitations}
The limitations of our work can be summarized as follows: Firstly, due to the scarcity of well-defined real-world constraint satisfaction problems, we relied on puzzle-like data for evaluating LLMs' reflective reasoning capabilities. Secondly, the inherent complexity and verbosity of LLM-generated responses to these complex reasoning tasks posed challenges for more fine-grained analysis. We only analyze several typical phenomena of current leading models rather than conducting a more detailed analysis of specific reflective reasoning processes.

\section*{Acknowledgments}
We thank our colleagues Qianlong Du, Fuwei Cui, Junhong Wu, Yangyifan Xu, Yupu Liang, Tianyu Peng, Wei Sun, and Tengxiao Xi for their insightful and constructive feedback. We thank Huaiguang Cai for assistance in conducting the evaluation of DeepSeek-R1. Furthermore, we thank all reviewers for their detailed reviews and valuable comments. This work is supported by the National Key R\&D Program of China No.2022ZD0160602 and the Strategic Priority Research Program of Chinese Academy of Sciences under Grant No.XDA04080400. This work is also supported by Zhongguancun Academy Project No.20240103.

\bibliography{custom}

\clearpage

\appendix

\section{Task Example}
\label{app_example}
In this section, we provide specific examples of each type of task to facilitate a better understanding of data samples in LR$^2$Bench.

\subsection{Crossword}
\begin{tcolorbox}[breakable, title=Crossword Example, colframe=blue!60]
\begin{CustomVerbatim}
Grid:
#   1   2   3   #
4   ?   ?   ?   5
6   ?   ?   ?   ?
7   ?   ?   ?   ?
8   ?   ?   ?   #

Clues:
Across:
1: "When They See Us" director DuVernay (3)
4: WNBA team based in Seattle (5)
6: Locations for orations (5)
7: Signed (5)
8: Lipton products (4)
Down:
1: Be penitent (5)
2: Black Russian ingredient (5)
3: First sign of the zodiac (5)
4: Barbecue rod (4)
5: Fuming (3)

Answer:
#   A   V   A   #
S   T   O   R   M
P   O   D   I   A
I   N   K   E   D
T   E   A   S   #
\end{CustomVerbatim}
\end{tcolorbox}

\subsection{Acrostic}
\begin{tcolorbox}[breakable, title=Acrostic Example, colframe=blue!60]
\begin{CustomVerbatim}
Grid:
1 2 3 4 # 5 6 7 # 8 9 10 11 # 12 13 14 # 15 16 17 18 19 20 21 22 # 23 24 25 26 27 28 # 29 30 31 32 33 34 # 35 # 36 37 38 39 40 41 ' 42 # 43 44 45 46 . 47 48 49 # 50 51 52 # 53 54 55 56 57 58 # 59 60 # 61 62 # 63 64 65 66 , 67 68 # 69 70 71 72 73 74 # 75 76 # 77 78 # 79 80 81 82 . 83 84 ' 85 # 86 # 87 88 89 90 # 91 92 93 94 95 96 .

Clues:
A
57 18 88 9 33 80 78
Gaynor hit \"I Will ____\" (1 word, 7 letters)
B
29 70 67 82 21
Wheat or milk beginning (1 word, 5 letters)
C
5 7 24 11 35 75 90
Extol (1 word, 7 letters)
D
30 50 10 91 46 14
Wintry projection (1 word, 6 letters)
E
52 26 73 58 94
Second-generation Japanese (1 word, 5 letters)
F
59 93 45 95 54 43 84 65 41 27
Standard (1 word, 10 letters)
G
92 72 60 4 37 66
Like some sweatshirts (1 word, 6 letters)
H
74 34 53 64 28 89 22
Scrambles a message (1 word, 7 letters)
I
36 25 2 61 86 23 83 19 81 31 47
Odds (1 word, 11 letters)
J
38 96 63 15 40 6 85
The Arctic and the Antarctic, e.g. (1 word, 7 letters)
K
71 8 49 87 20
Parisian eggs (1 word, 5 letters)
L
1 56 76 77 79 13
Peanut (1 word, 6 letters)
M
68 48 55 39
Aussie jumpers (1 word, 4 letters)
N
12 17 44 42 62
Strike ___ (what models do) (2 words, 5 letters)
O
16 51 69 32 3
Far from wimpy (1 word, 5 letters)

Answer:
"Good and evil are impulses buried within a person's soul. You can choose to be good, or choose to be evil. It's a free choice." - James Swain, The Program
\end{CustomVerbatim}
\end{tcolorbox}

\subsection{Logic Puzzle}
\begin{tcolorbox}[breakable, title=Logic Puzzle Example, colframe=blue!60]
\begin{CustomVerbatim}
Variables:
{
    "anniversaries": [
        "January 28",
        "March 6",
        "November 2",
        "November 23"
    ],
    "first names": [
        "Asher",
        "Bradley",
        "Kayla",
        "Malia"
    ],
    "conveyances": [
        "10-speed bike",
        "mountain bike",
        "segway",
        "skateboard"
    ],
    "actors": [
        "Liam Neeson",
        "Morgan Freeman",
        "Robert Duvall",
        "Tom Cruise"
    ]
}

Clues:
1. The commuter who rides a 10-speed bike is not Bradley.
2. Kayla isn't related to Robert Duvall and doesn't ride a segway.
3. Robert Duvall's cousin has an anniversary after the commuter who rides a segway.
4. Morgan Freeman's cousin is not Asher and doesn't ride a segway.
5. Liam Neeson's cousin is Malia.
6. Liam Neeson's cousin uses a mountain bike to get to work.
7. Of Malia and the commuter who rides a 10-speed bike, one has an anniversary on January 28 and the other has an anniversary on November 23.
8. Either the person whose anniversary is on March 6 or the person whose anniversary is on November 2 is Morgan Freeman's cousin.
9. Tom Cruise's cousin has an anniversary after the commuter who rides a skateboard.

Answer:
[
    {
        "anniversaries": "January 28",
        "first names": "Malia",
        "conveyances": "Mountain bike",
        "actors": "Liam Neeson"
    },
    {
        "anniversaries": "March 6",
        "first names": "Kayla",
        "conveyances": "Skateboard",
        "actors": "Morgan Freeman"
    },
    {
        "anniversaries": "November 2",
        "first names": "Bradley",
        "conveyances": "Segway",
        "actors": "Tom Cruise"
    },
    {
        "anniversaries": "November 23",
        "first names": "Asher",
        "conveyances": "10-speed bike",
        "actors": "Robert Duvall"
    }
]
\end{CustomVerbatim}
\end{tcolorbox}

\subsection{Cryptogram}
\begin{tcolorbox}[breakable, title=Cryptogram Example, colframe=blue!60]
\begin{CustomVerbatim}
Cryptogram:
VYQ'O CY HGYWQV THMFQC OKB LYGUV YLBT MYW H UFJFQC. OKB LYGUV YLBT MYW QYOKFQC. FO LHT KBGB RFGTO. - Mark Twain

Answer:
Don't go around saying the world owes you a living. The world owes you nothing. It was here first. - Mark Twain
\end{CustomVerbatim}
\end{tcolorbox}

\subsection{Sudoku}
\begin{tcolorbox}[breakable, title=Sudoku Example, colframe=blue!60]
\begin{CustomVerbatim}
Sudoku:
[[4, 0, 0, 0],
 [0, 3, 2, 0],
 [0, 4, 1, 0],
 [0, 0, 0, 2]]

Answer:
[[4, 2, 3, 1],
 [1, 3, 2, 4],
 [2, 4, 1, 3],
 [3, 1, 4, 2]]
\end{CustomVerbatim}
\end{tcolorbox}

\subsection{Drop Quote}
\begin{tcolorbox}[breakable, title=Drop Quote Example, colframe=blue!60]
\begin{CustomVerbatim}
Grid:
1  2  #  4  5  6  7  8  9   #   11 12 13 14 15
#  2  3  4  5  6  7  8  9  10  11  #  13 14 15
1  2  #  4  5   '  7  #  9  10  11 12  .    #   #

Column 1: I I
Column 2: I L T
Column 3: M
Column 4: A I P
Column 5: L O T
Column 6: S W
Column 7: A S S
Column 8: I Y
Column 9: B D S
Column 10: L O
Column 11: E N S
Column 12: E E
Column 13: E U
Column 14: M N
Column 15: S T

Answer:
It always seems impossible until it's done. - Nelson Mandela
\end{CustomVerbatim}
\end{tcolorbox}

\section{Prompt Templates}
\label{app_prompt}
We provide detailed prompt templates for each type of task in LR$^2$Bench, including task descriptions, examples, and solution guidelines.
\subsection{Crossword}
\begin{tcolorbox}[breakable, title=Prompt for Crossword]
\small
\textbf{[Task Description]} \\

A Crossword puzzle is a word game that consists of a grid, with clues given for words that fit into the grid both across (horizontally) and down (vertically). Your goal is to fill in the grid with words based on the clues provided. Here's a detailed explanation of how the game works: \\

1. Understand the Grid Layout \\
The grid is made up of numbers, hashtags ("\#"), and question marks ("?"). \\

Hashtag ("\#") acts as a separator between words. \\

The number represents the starting points of Across and Down words. \\

Question mark ("?") represents part of words but don't start a new word. \\

2. Read the Clues \\
Clues are provided for each word to be filled into the grid, split into two categories: Across clues (these are for words that go horizontally in the grid) and Down clues (these are for words that go vertically in the grid). \\

The number in brackets after the clue indicates the length of the word. \\

Clues are often short definitions, synonyms, or phrases related to the word. \\

Some clues may involve wordplay, anagrams, or puns, depending on the puzzle's difficulty and style. \\

3. Solve the Puzzle \\
Think of words that fit the clue and match the number of letters specified. \\

For example, if a clue says "Animal that barks (3)", you might guess "DOG" because it has 3 letters. When getting the answer, double-check that it fits the clue, both in meaning and the number of letters. If you make a mistake, just try again! \\

As you fill in answers, they will help you solve other clues, since words intersect and share letters in the grid. This cross-checking mechanism helps in verifying correct answers. \\

\textbf{[Example 1]}

\begin{CustomVerbatim}
Here is a Crossword puzzle:
GRID:
# 1 2 3 4
5 ? ? ? ?
6 ? ? ? ?
7 ? ? ? ?
# 8 ? ? #

CLUES:
Across:
1 "More ___ you'll ever know" (4)
5 Divine favor, in Christianity (5)
6 Sisters who might be sitters (5)
7 Portly-sounding beer (5)
8 Cool and stylish (3)

Down:
1 "The ___ will set you free. But not until it is finished with you": David Foster Wallace (5)
2 Capital of Vietnam (5)
3 Misbehave (5)
4 Prepare for a new baby, say (4)
5 Fuel for some stoves (3)
\end{CustomVerbatim}

\textbf{[Solution 1]}

\begin{CustomVerbatim}
Let's start solving the Crossword puzzle step by step.

Across Clues:

<1-Across>
1-Across: "More ___ you'll ever know" (4)

The clue is a reference to a popular phrase, "More than you'll ever know," suggesting the answer is "than".
Answer: THAN.
</1-Across>

<5-Across>
5-Across: Divine favor, in Christianity (5)

The clue refers to a concept of God's blessing or grace, and the answer is "grace".
Answer: GRACE.
</5-Across>

... ...

Here is the final grid after finishing all the clues:
<Answer>
{"1-Across": "THAN", "5-Across": "GRACE", "6-Across": "AUNTS", "7-Across": "STOUT", "8-Across": "HIP", "1-Down": "TRUTH", "2-Down": "HANOI", "3-Down": "ACTUP", "4-Down": "NEST", "5-Down": "GAS"}
</Answer>
\end{CustomVerbatim}

\textbf{[Example 2]} \\

<Example 2> \\

\textbf{[Solution 2]} \\

<Solution 2> \\

\textbf{[Solution Guideline]} \\

Please solve the Crossword Puzzle according to the provided rules. Please also follow the requests below to present your analysis and solutions: \\

1. Analyze each clue carefully to understand its meaning and potential word associations. Be open to the possibility of wordplay or puns that might lead to unexpected solutions. The filled-in words will help you infer the answers to the subsequent clues. \\

2. Provide the final answer for each clue. The final answer should be presented after "Answer:". \\

3. Please wrap all the analysis of each clue with <n-Across> and </n-Across> or <n-Down> and </n-Down> tags, where n is the clue number. For example, the Across clue 1 should be wrapped with <1-Across> and </1-Across> tags. The Down clue 1 should be wrapped with <1-Down> and </1-Down> tags, and so on. \\

4. After solving all the clues, please summarize all the answer words in following json format and warp them with <Answer> and </Answer> tags: \\
<Answer> \\
\{"1-Across": "ANSWER1", "2-Across": "ANSWER2", ... "1-Down": "ANSWER3", "2-Down": "ANSWER4", ...\} \\
</Answer> \\

5. Please generate your response without redundant and repeating content. \\

\textbf{[Question]} \\

<Question>
\end{tcolorbox}

\subsection{Acrostic}
\begin{tcolorbox}[breakable, title=Prompt for Acrostic]
\small
\textbf{[Task Description]} \\

An Acrostic puzzle is a word game that consists of two main parts: a grid and a set of clues. The objective is to fill in the grid with letters from the answers to the clues, forming a hidden quotation or sentence. Here's a detailed explanation of how the game works: \\

1. Understand the Grid Layout \\
The grid is made up of numbers, hashtags ("\#"), and punctuations. Hashtag ("\#") acts as space separator between words. \\

Each number corresponds to a specific letter in a word. The hidden quotation or sentence is formed by filling in the grid with the correct letters. \\

2. Read the Clues \\
Each clue is made up of a string of numbers and a clue text. \\

The clue text is usually a short definition, synonym or phrase related to the answer word or phrase. \\

The number of words and letters of the answer are given in brackets after the clue text. \\

The string of numbers provided indicates the specific position of each letter of the answer in the grid. \\

3. Solve the Puzzle \\
Think of answers that fit the clue and match the number of words and letters specified. \\

For example, if a clue says "Animal that barks (1 word, 3 letters)", you might guess "DOG" because it has 3 letters. When getting the answer, double-check that it fits the clue, both in meaning and the number of letters. If you make a mistake, just try again! \\

As you fill in the letters of answers, these letters can be part of the hidden word in the grid, thus providing hints for the left letters of this word. This mechanism can help both verify your answers and solve difficult clues. \\

For example, if the puzzle looks like this: "... \# 5 6 7 \# ..." and you have already got \{5: 'A', 7: 'D'\}, you can guess that the word is "AND" and get \{6: 'N'\} without its corresponding clue. Don't be afraid to skip a tough clue and come back to it later. \\

As you fill in the grid, you will reveal part of the hidden words. Use this information to help solve the remaining clues. \\

\textbf{[Example 1]}

\begin{CustomVerbatim}
Here is an Acrostic puzzle:
GRID:
1 2 3 4 5 # 6 7 8 9 10 # 11 12 # 13 14 15 # 16 17 18 19 20 # 21 22 23 24 25 26 27 28 29 , 30 31 32 # 33 34 35 # 36 37 38 # 39 40 41 42 43 # 44 45 46 47 # 48 49 50 # 51 52 53 54 55 56 57 # 58 59 60 61 # 62 63 64 65 66 67 68 69 70 # 71 72 73 74 # 75 # 76 77 78 79 # 80 81 82 # 83 84 85 # 86 87 88 89 90 91 92 .

CLUES:
A
58 86 92 3 60 49 57 46 70 40
Cosmetics magnate (3 words, 10 letters)
B
52 34 44 54 84 67 11 36 25 18 62
Nonphysical assets (1 word, 11 letters)
C
76 45 35 88 56 81 43 41 68 71
Religious schools of opinion (1 word, 10 letters)
D 
48 37 16 24 82 85 30 63 73
Cancer stick (1 word, 9 letters)
E
10 32 2 74 22 23 75 79 39 5 64
Establish a home base (1 word, 11 letters)
F
59 9 14 69 83 53 28
"____ is only a bench covered in velvet" (quote by Napoleon) (2 words, 7 letters)
G
55 26 38 31 20 21 87 91 6
___ up (became less serious) (1 word, 9 letters)
H
4 17 61 65 13 90 77 29
Rest (1 word, 8 letters)
I
33 80 51 15 66 27 7 12
Infractions (1 word, 8 letters)
J
1 19 50 72 8 89 47 78 42
Whisky-vermouth cocktail (1 word, 9 letters)
\end{CustomVerbatim}

\textbf{[Solution 1]}

\begin{CustomVerbatim}
Let's start solving the Acrostic puzzle step by step.

<A-CLUE>
A
58 86 92 3 60 49 57 46 70 40
Cosmetics magnate (3 words, 10 letters)

The clue "Cosmetics magnate" succinctly encapsulates Mary Kay Ash's pioneering role in establishing a direct sales empire in the beauty industry.
Answer: MARYKAYASH.
Mapping: {'58': 'M', '86': 'A', '92': 'R', '3': 'Y', '60': 'K', '49': 'A', '57': 'Y', '46': 'A', '70': 'S', '40': 'H'}
</A-CLUE>

<B-CLUE>
B
52 34 44 54 84 67 11 36 25 18 62
Nonphysical assets (1 word, 11 letters)

"Nonphysical assets" aptly refers to intangibles, highlighting assets like intellectual property and goodwill that lack physical form but possess significant business value.
Answer: INTANGIBLES.
Mapping: {'52': 'I', '34': 'N', '44': 'T', '54': 'A', '84': 'N', '67': 'G', '11': 'I', '36': 'B', '25': 'L', '18': 'E', '62': 'S'}
</B-CLUE>

... ...

Now let's gather all the mappings in order:
<Answer>
{'1': 'M', '2': 'A', '3': 'Y', '4': 'B', '5': 'E', '6': 'D', '7': 'E', '8': 'A', '9': 'T', '10': 'H', '11': 'I', '12': 'S', '13': 'T', '14': 'H', '15': 'E', '16': 'G', '17': 'R', '18': 'E', '19': 'A', '20': 'T', '21': 'E', '22': 'Q', '23': 'U', '24': 'A', '25': 'L', '26': 'I', '27': 'S', '28': 'E', '29': 'R', '30': 'T', '31': 'H', '32': 'E', '33': 'O', '34': 'N', '35': 'E', '36': 'B', '37': 'I', '38': 'G', '39': 'T', '40': 'H', '41': 'I', '42': 'N', '43': 'G', '44': 'T', '45': 'H', '46': 'A', '47': 'T', '48': 'C', '49': 'A', '50': 'N', '51': 'F', '52': 'I', '53': 'N', '54': 'A', '55': 'L', '56': 'L', '57': 'Y', '58': 'M', '59': 'A', '60': 'K', '61': 'E', '62': 'S', '63': 'T', '64': 'R', '65': 'A', '66': 'N', '67': 'G', '68': 'E', '69': 'R', '70': 'S', '71': 'S', '72': 'H', '73': 'E', '74': 'D', '75': 'A', '76': 'T', '77': 'E', '78': 'A', '79': 'R', '80': 'F', '81': 'O', '82': 'R', '83': 'O', '84': 'N', '85': 'E', '86': 'A', '87': 'N', '88': 'O', '89': 'T', '90': 'H', '91': 'E', '92': 'R'}
</Answer>
\end{CustomVerbatim}

\textbf{[Example 2]} \\

<Example 2> \\

\textbf{[Solution 2]} \\

<Solution 2> \\

\textbf{[Solution Guideline]} \\

Please solve the Acrostic Puzzle according to the provided rules. Please also follow the requests below to present your analysis and solutions: \\

1. Analyze the clue carefully to understand its meaning and potential word associations. Be open to the possibility of wordplay or puns that might lead to unexpected solutions. The filled-in words will help you infer the answers to the subsequent clues. \\

2. Provide the final answer for each clue. The final answer should be presented after "Answer:". \\

3. Create a python dictionary mapping in a single line that links the number positions in the clues to the corresponding letters in the final answer. Note that the blank spaces and punctuation should be omitted in the mapping. The python dictionary should be presented after "Mapping:". \\

4. Please wrap all the analysis of each clue with <n-CLUE> and </n-CLUE> tags, where n is the label of the clue. For example, the first clue should be wrapped with <A-CLUE> and </A-CLUE> tags. The second clue should be wrapped with <B-CLUE> and </B-CLUE> tags, and so on. \\

5. After solving all the clues, please gather all the mappings and place them in order between the <Answer> and </Answer> tags. The gathered mapping should look like \{'1': 'LETTER1', '2': 'LETTER2', ...\}. \\

6. Please generate your response without redundant and repeating content. \\

\textbf{[Question]} \\

<Question>
\end{tcolorbox}

\subsection{Logic Puzzle}
\begin{tcolorbox}[breakable, title=Prompt for Logic Puzzle]
\small
\textbf{[Task Description]} \\

Logic puzzles require the solver to deduce the relationships between different people, places and things based on a limited number of clues given in the puzzle. Remember: every item belongs to one and only one person, no item will ever be shared. Using only the clues provided and simple deductive logic and reasoning. \\

\textbf{[Example 1]}

\begin{CustomVerbatim}
Here is a logic puzzle:
{
    "variables": {
        "names": [
            "Fiona",
            "Tobias",
            "Leanne"
        ],
        "main dishes": [
            "Beef Wellington",
            "Gado Gado",
            "Shrimp Ceviche"
        ],
        "desserts": [
            "Creme Brulee",
            "Raspberry Tiramisu",
            "French Apple Cake"
        ]
    },
    "clues": {
        "1": "The three people are Tobias, the one who brought the Gado Gado, and the one who brought the Raspberry Tiramisu.",
        "2": "Fiona did not bring the Beef Wellington.",
        "3": "The Shrimp Ceviche was brought by either Tobias or the person who brought the Creme Brulee.",
    }
}

Please complete the following blank based on your final solution:
[
    {
        "names": "Fiona",
        "main dishes": "____",
        "desserts": "____",
    },
    {
        "names": "Tobias",
        "main dishes": "____",
        "desserts": "____",
    },
    {
        "names": "Leanne",
        "main dishes": "____",
        "desserts": "____",
    }
]
\end{CustomVerbatim}

\textbf{[Solution 1]}

\begin{CustomVerbatim}
Let's start solving the logic puzzle step by step.

Participants and Dishes
Names: Fiona, Tobias, Leanne
Main Dishes: Beef Wellington, Gado Gado, Shrimp Ceviche
Desserts: Crème Brûlée, Raspberry Tiramisu, French Apple Cake


Interpreting Clue 1:
    Roles:
        Tobias: Did not bring Gado Gado or Raspberry Tiramisu.
        Gado Gado Bringer: Not Tobias.
        Raspberry Tiramisu Bringer: Not Tobias.

    Conclusion:
        Fiona and Leanne must be the ones who brought Gado Gado and Raspberry Tiramisu, respectively (or vice versa).

... ...

Final Answer in JSON Format:
<Answer>
[
    {
        "names": "Fiona",
        "main dishes": "Gado Gado",
        "desserts": "Creme Brulee",
    },
    {
        "names": "Tobias",
        "main dishes": "Shrimp Ceviche",
        "desserts": "French Apple Cake",
    },
    {
        "names": "Leanne",
        "main dishes": "Beef Wellington",
        "desserts": "Raspberry Tiramisu",
    }
]
</Answer>
\end{CustomVerbatim}

\textbf{[Example 2]} \\

<Example 2> \\

\textbf{[Solution 2]} \\

<Solution 2> \\

\textbf{[Solution Guideline]} \\

Please solve the Logic Puzzle according to the provided rules. Please also follow the requests below to present your analysis and solutions: \\

1. Read and understand each clue in the context of the puzzle. Apply each clue one by one to deduce the correct arrangement of different variables. \\

2. Use logical reasoning to figure out the relationships between the variables based on the clues provided. \\

3. After solving the puzzle, present your final solution using JSON Format and wrap it with the <Answer> and </Answer> tags. For example:

\begin{CustomVerbatim}
<Answer>
[
    {
        "Variable1": "Value1",
        "Variable2": "Value2",
        ...
    },
    {
        "Variable1": "Value3",
        "Variable2": "Value4",
        ...
    },
    ...
]
</Answer>
\end{CustomVerbatim}

4. Please generate your response without redundant and repeating content. \\

\textbf{[Question]} \\

<Question>
\end{tcolorbox}

\subsection{Cryptogram}
\begin{tcolorbox}[breakable, title=Prompt for Cryptogram]
\small
\textbf{[Task Description]} \\

A cryptogram is a type of puzzle that consists of a short piece of encrypted text. You will decode a cryptogram that uses a simple substitution cipher, where each letter in the message has been substituted with another letter. The goal is to decode the hidden quote along with its author's name. You may use the author's name to help reduce the difficulty of the decoding process by focusing on common words, letter patterns, and stylistic cues from the author. \\

\textbf{[Example 1]} \\

Think step-by-step to decipher the cryptogram: \\
VOBB XSDO QK IOFFOU FGHD VOBB KHQX. \\
— Benjamin Franklin \\

\textbf{[Solution 1]} \\

Let's start solving the cryptogram puzzle step by step. \\

Step 1: Use the Author's Name as a Clue Benjamin Franklin is known for his concise and wise sayings. Recognizing his style helps anticipate common structures and themes, such as practical advice or moral lessons. \\

Step 2: Analyze the Cryptogram The cryptogram is: \\
VOBB XSDO QK IOFFOU FGHD VOBB KHQX. -Benjamin Franklin \\

Notable features: \\
"VOBB" appears twice, suggesting a common word with double letters. \\
The structure of the sentence hints at a proverb or a well-known quote. \\

... ... \\

\begin{CustomVerbatim}
Mapping:
<Mapping>
{ "V": "W", "O": "E", "B": "L", "X": "D", "S": "O", "D": "N", "Q": "I", "K": "S", "I": "B", "F": "T", "U": "R", "G": "H", "H": "A"}
</Mapping>

Answer:
<Answer>
Well done is better than well said.
-Benjamin Franklin
</Answer>
\end{CustomVerbatim}

\textbf{[Example 2]} \\

<Example 2> \\

\textbf{[Solution 2]} \\

<Solution 2> \\

\textbf{[Solution Guideline]} \\

Please follow these steps to solve the Cryptogram: \\

1. Use the author's name as a clue: Knowing the author's name can help you predict common words or letter combinations typical for this author. For instance, if the author is "Shakespeare," you might anticipate archaic or common Shakespearean phrases (like "thou," "thee," etc.). \\

2. Analyze the cryptogram: Look at the frequency of letters and common letter patterns, such as double letters or common suffixes and prefixes. Focus on the parts of the cryptogram that seem to match the author's typical writing style or famous phrases. \\

3. Map common words: If you recognize a word in the cryptogram that matches the author's typical vocabulary, substitute letters based on that. \\

4. Make educated guesses: Use common English words (such as "the," "and," "of," etc.) and letter pairs (like "th," "he," "in", etc.) to identify possible substitutions. If one assumption doesn't work, try another. \\

5. Verification: After generating the decoded message, check if the quote and the author's name make logical sense. If needed, revisit the assumptions and adjust the letter mappings. \\

6. Please provide step by step analysis and create a python dictionary mapping of the fully substitutioin wrapping it between <Mapping> and </Mapping> tag. \\

7. Please provide the final decoded quote and wrap it between <Answer> and </Answer> tag. \\

8. Please generate your response without redundant and repeating content. \\

\textbf{[Question]} \\

<Question>
\end{tcolorbox}

\subsection{Sudoku}
\begin{tcolorbox}[breakable, title=Prompt for Sudoku]
\small
\textbf{[Task Description]} \\

Sudoku is a logic-based, combinatorial number-placement puzzle. The puzzle consists of an $n^2 \times n^2$ grid, partially filled with numbers from 1 to $n^2$ and 0 (empty cells). The objective is to fill the grid so that each row, each column, and each $n \times n$ subgrid must contain every number from 1 to $n^2$ exactly once. \\

\textbf{[Example 1]}

\begin{CustomVerbatim}
Here is a 4 x 4 sudoku puzzle:
[[2, 0, 0, 0],
 [0, 0, 3, 0],
 [0, 4, 0, 0],
 [0, 0, 0, 1]]
\end{CustomVerbatim}

\textbf{[Solution 1]}

\begin{CustomVerbatim}
Let's start solving the sudoku puzzle step by step.

Step1: Analyze Empty Cell in Row 1
Cell (1,2):
    Eliminated Numbers: 2 (already in row), 4 (already in column)
    Possible Numbers: 1, 3
Cell (1,3):
    Eliminated Numbers: 2 (already in row), 3 (already in column)
    Possible Numbers: 1, 4
Cell (1,4):
    Eliminated Numbers: 2 (already in row), 1 (already in column), 3 (already in subgrid)
    Only possible number: 4

... ...

Step 5: Final Answer
<Answer>
[[2, 3, 1, 4],
 [4, 1, 3, 2],
 [1, 4, 2, 3],
 [3, 2, 4, 1]]
</Answer>
\end{CustomVerbatim}

\textbf{[Example 2]} \\

<Example 2> \\

\textbf{[Solution 2]} \\

<Solution 2> \\

\textbf{[Solution Guideline]} \\

Please solve the Sudoku Puzzle according to the provided rules. Please also follow the requests below to present your analysis and solutions: \\

1. For each empty cell, try to deduce which numbers are allowed based on the existing numbers in its row, column, and subgrid. \\

2. If you reach a point where no obvious choices are available, you may need to backtrack. Try filling in a number and see if it leads to a valid solution. If it causes a contradiction, backtrack to a previous decision and try a different path. \\

3. Please provide step by step analysis and present the final answer as the same json format of the input grid. Wrapping the final answer with <Answer> and </Answer> tags. \\

4. Please generate your response without redundant and repeating content. \\

\textbf{[Question]} \\

<Question>
\end{tcolorbox}

\subsection{Drop Quote}
\begin{tcolorbox}[breakable, title=Prompt for Drop Quote]
\small
\textbf{[Task Description]} \\

A Drop Quote Puzzle consists of two parts: a grid and a set of given letters that "drop" into specific columns of the grid. The goal is to fill in the grid with a meaningful quote or statement. \\

Key Features: \\
1. The grid contains numbers, "\#" symbols (representing spaces between words), and already placed punctuation. \\

2. Words may break across lines, starting at the end of one line and continuing to the beginning of the next. Words end only at "\#" blocks or the end of the puzzle. \\

3. The letters are organized by column number and are given in alphabetical order. These letters need to be placed into the grid. \\

4. Each letter drops into a specific column, and your task is to identify the correct placement for each letter based on the grid structure. Do not simply fill the columns in the order they are listed. Instead, consider where each letter fits contextually within the puzzle to form words that make sense in the final quote or statement. \\

Rules: \\
1. The "\#" symbols indicate where spaces between words should be. \\

2. Column labels (e.g., Column 1: O T) indicate the letters that will drop into the grid for Column 1, listed in alphabetical order. For example, in Column 1, the letters "O" and "T" must be placed in the appropriate grid positions. \\

3. Fill the grid with the letters in such a way that they form a meaningful statement or quotation. \\

\textbf{[Example 1]}

\begin{CustomVerbatim}
Here is a drop quote puzzle:
[['1', '2', '3', '4', '#', '6', '7'],
 ['#', '2', '3', '4', '5', '6', '.']]

Column 1: T
Column 2: I M
Column 3: M O
Column 4: E N
Column 5: E
Column 6: I Y
Column 7: S
\end{CustomVerbatim}

\textbf{[Solution 1]}

\begin{CustomVerbatim}
Let's start solving the sudoku puzzle step by step.

We first focus on Columns with single letters:
    "T" for Row 1, Column 1
    "E" for Row 2, Column 5
    "S" for Row 1, Column 7

Then, we check the word ['6', '7'] in Row 1:
    The candidates for Column 6 are "I" and "Y", and for Column 7, it is "S". The only word that fits is "IS". So, place "I" in Row 1, Column 6 and "Y" in Row 2, Column 6.

Let's check the current grid:
[['T', '2', '3', '4', '#', 'I', 'S'],
 ['#', '2', '3', '4', 'E', 'Y', '.']]

... ...

Here is the final answer gird:
<Answer>
[['T', 'I', 'M', 'E', '#', 'I', 'S'],
 ['#', 'M', 'O', 'N', 'E', 'Y', '.']]
</Answer>
\end{CustomVerbatim}

\textbf{[Example 2]} \\

<Example 2> \\

\textbf{[Solution 2]} \\

<Solution 2> \\

\textbf{[Solution Guideline]} \\

Please solve the Drop Quote Puzzle according to the provided rules. Please also follow the requests below to present your analysis and solutions: \\

1. Isolated boxes are usually, but not always, the words A or I. Similarly, when you have a 2-letter word, see what words you can make and then see what letters that leaves you over with for other lines. The most common 2-letter words are: OF, TO, IN, IT, IS, BE, AS, AT, SO, WE, HE, BY, OR, ON, DO, IF, ME, MY, UP, AN, GO, NO, US, AM. \\

2. The letters in a column with fewer letters can be placed quickly by noticing which letters must be consonants and vowels, or by eliminating the possibility of a letter appearing in a certain spot. \\

3. If you have an uncommon letter such as a J, K, Q, X, or Z, think about what letters might go before or after it. Similarly, look for common consonant pairs such as TH, ND, NT, ST, as well as doubled letters. \\

4. Be alert for common prefixes such as DE-, MIS-, RE-, and UN-, as well as common suffixes such as -ABLE, -ED, -ING, -LY, -NESS, and -TION. \\

5. Every time you fill in a letter, it means that you've eliminated some letters that can go elsewhere. Frequently, this means that you can immediately fill in some additional letters. \\

6. Please provide step by step analysis and present the final answer as the same json format of the input grid. Wrapping the final answer in <Answer> and </Answer> tags. \\

7. Please generate your response without redundant and repeating content. \\

\textbf{[Question]} \\

<Question>
\end{tcolorbox}

\section{Annotation}
\label{app_annotation}
All tasks with LR$^2$Bench were constructed by four annotators, all pursuing a Master's or PhD in Computer Science. Here is the task-specific allocation:
\paragraph{Crossword \& Acrostic (250 samples, 1 annotator)} Converting web-crawled content (grids, clues, and answers) into structured text formats; Designing instructions and few-shot examples.
\paragraph{Logic Puzzle (200 samples, 1 annotator)} Converting PDF data (variables, clues, and answers) into structured text formats; Designing instructions and few-shot examples.
\paragraph{Sudoku (200 samples, 1 annotator)} Converting PDF data (grids and answers) into structured text formats; Verifying Sudoku correctness with Python scripts; Designing instructions and few-shot examples.
\paragraph{Cryptogram \& Drop Quote (200 samples, 1 annotator)} Converting PDF data (encrypted quotations, grids, and answers) into structured text formats; Creating Drop Quote with common quotations; Designing instructions and few-shot examples.
Moreover, 50 samples per task were double-checked by different annotators, with the inter-annotator agreement (IAA) exceeding 95\%.

\section{Subtask Definition}
\label{app_subtask}
The subtasks of each type of task is defined as follows:
\paragraph{Crossword} Each word corresponding to a given clue (e.g., The answer to '1-Across' is the word 'THAN').
\paragraph{Acrostic} Each letter's placement in the grid as derived from the clue answers (e.g., The letter in the first position of the hidden quote is 'M').
\paragraph{Logic Puzzle} Each specific attribute assignment that needs to be deduced (e.g., Person A owns a red car).
\paragraph{Cryptogram} Each correct mapping between an encrypted letter and its original letter (e.g., The letter 'V' in the cryptogram corresponds to the letter 'M').
\paragraph{Sudoku} Each individual empty cell that needs to be filled with the correct digit (e.g., The cell at Row 1, Column 1 should contain the digit '1').
\paragraph{Drop Quote} Each individual cell in the grid that needs to be filled with the correct letter (e.g., The cell at Row 1, Column 1 should contain the letter 'T').

\section{Models}
\label{app_models}
As shown in Table \ref{models}, we list the key details for eleven open-source models and five closed-source models in our evaluation.

\section{Few-shot Examples}
\label{app_shot}

\begin{table}[!h]
    \centering
    \resizebox{1.0\columnwidth}{!}{
    \begin{tabular}{l|ccccccc}
        \toprule
        \textbf{Model} & \textbf{Few-shot} & \textbf{CR} & \textbf{S-Acc} & \textbf{EM} & \textbf{PM-0.5} & \textbf{Redundancy} & \textbf{\# Tokens} \\
        \midrule
        \multirow{2}{*}{Llama-3.1-8B} & Manually & 57.0 & 16.0 & 0.0 & 8.0 & 44.5 & 1,293 \\
        & DeepSeek-R1 & 9.5 & 3.3 & 0.0 & 2.5 & 91.0 & 1,731 \\
        \multirow{2}{*}{Qwen-2.5-7B} & Manually & 96.5 & 25.8 & 0.0 & 8.5 & 4.0 & 1,396 \\
        & DeepSeek-R1 & 96.5 & 23.3 & 0.0 & 6.0 & 3.0 & 1,596 \\
        \bottomrule
    \end{tabular}}    
    \caption{Performance of the Logic Puzzle task with different few-shot examples.}
    \label{tab:few_shot}
\end{table}

The few-shot provided to LLMs are simpler than the tested ones. Our intention is primarily to illustrate the task format and desired output structure, ensuring the models understand how to present their solutions for evaluation. Moreover, we aim to assess the inherent reflective reasoning capabilities of the models themselves, rather than explicitly prompting or guiding them toward exhibiting reflection through the few-shot examples. While we acknowledge that carefully crafted prompts can sometimes elicit improved performance in conventional models, our focus in LR$^2$Bench is on evaluating the models' intrinsic capacity for long-chain reflective reasoning without extensive external guidance within the prompt itself.

In Table \ref{tab:few_shot}, we provide the results of the Logic Puzzle task with few-shot examples constructed by DeepSeek-R1. We observed that conventional models struggle to benefit from few-shot examples with the reflection process. They only tend to mimic the reflection format (the increasing generated tokens), showing inferior performance due to the lack of authentic reflective mechanisms. Specifically, Llama-3.1-8B-Instruct falls into redundancy when trying to perform reflection as the few-shot examples, leading to significant performance degradation.

\section{Task Difficulty}
\label{app_difficulty}

\begin{table}[!h]
\centering
\resizebox{1.0\columnwidth}{!}{\begin{tabular}{llc}
\toprule
\textbf{Task} &  
\textbf{Difficulty} &

\begin{tabular}[c]{@{}c@{}}\hspace{0.5em} \textbf{Solution Space} \hspace{0.5em} \\ \textbf{Upper Bound} \end{tabular} \\
\midrule
\multirow{3}{*}{Crossword} & $5\times5$ & $26\times5\times5$ \\
& $10\times10$ & $26\times10\times10$ \\
& $15\times15$ & $26\times15\times15$ \\
\midrule
\multirow{2}{*}{Acrostic} & Easy & $26\times110$ \\
& Hard & $26\times214$ \\
\midrule
\multirow{4}{*}{Logic Puzzle} & $4\times4$ & ${(4!)}^3$ \\
& $4\times5$ & ${(4!)}^4$ \\
& $4\times6$ & ${(4!)}^5$ \\
& $4\times7$ & ${(4!)}^6$ \\
\midrule
\multirow{2}{*}{Cryptogram} & Easy & $26!$ \\
& Hard & $26!$ \\
\midrule
\multirow{4}{*}{Sudoku} & $4\times4$, Easy & $4^{16}$ \\
& $4\times4$, Hard & $4^{16}$ \\
& $9\times9$, Easy & $9^{81}$ \\
& $9\times9$, Hard & $9^{81}$ \\
\midrule
\multirow{2}{*}{Drop Quote} & Easy & $15\times3!$ \\
& Hard & $20\times5!$ \\
\bottomrule
\end{tabular}}
\caption{The solution space upper bound of each type of task.}
\label{space}
\end{table}

% \begin{table}[t]
%     \centering
%     \resizebox{1.0\columnwidth}{!}{
%     \begin{tabular}{lcc}
%         \toprule
%         \textbf{Task} & \textbf{Difficulty} & \textbf{Solution Space Upper Bound}\\
%         \midrule
%         \multirow{3}{*}{Crossword} & $5\times5$ & $26\times5\times5$ \\
%         & $10\times10$ & $26\times10\times10$ \\
%         & $15\times15$ & $26\times15\times15$ \\
%         \midrule
%         \multirow{2}{*}{Acrostic} & Easy & $26\times110$ \\
%         & Hard & $26\times214$ \\
%         \midrule
%         \multirow{4}{*}{Logic Puzzle} & $4\times4$ & ${4!}^3$ \\
%         & $4\times5$ & ${4!}^4$ \\
%         & $4\times6$ & ${4!}^5$ \\
%         & $4\times7$ & ${4!}^6$ \\
%         \midrule
%         \multirow{2}{*}{Cryptogram} & Easy & $26!$ \\
%         & Hard & $26!$ \\
%         \midrule
%         \multirow{4}{*}{Sudoku} & $4\times4$, Easy & ${4!}^3$ \\
%         & $4\times4$, Hard & ${4!}^4$ \\
%         & $9\times9$, Easy & ${4!}^5$ \\
%         & $9\times9$, Hard & ${4!}^6$ \\
%         \midrule

%     \end{tabular}}    
%     \caption{.}
%     \label{tab:main_exp_sum}
% \end{table}

Table \ref{space} analyzes the difficulty of each type of task through the size of solution space. We approximately calculate the solution space by multiplying the number of elements to be filled (e.g., grid size for Crossword) by the number of possible answers for each element (e.g., 26 letters for Crossword). Detailed performance across varying difficulty levels for each task is presented in Tables \ref{level_crossword}, \ref{level_acrostic_cryptogram}, \ref{level_logic}, \ref{level_sudoku}, and \ref{level_drop}. Our findings indicate that the CR score does not significantly decrease with increasing task difficulty. This suggests that the long-chain generation capability is likely an inherent property of these models, rather than being strongly task-dependent. However, as the solution space expands with increasing difficulty, we observe a consistent performance decline across all models. In the most challenging settings, even the top-performing model achieves an EM score of less than 10\%, highlighting the difficulty of our benchmark for evaluating LLMs' reflective reasoning capabilities.

\section{Response Cases}
\label{app_case}

\subsection{Lack of Reflective Reasoning}
\label{lack}
The following example shows that GPT-4o fails to solve the logic problem due to the lack of reflective reasoning capabilities. The model only performs a single turn of deduction of all the clues and outputs the final answers with insufficient evidence.

\begin{tcolorbox}[breakable, colback=blue!8, title=Example 1: Response of GPT-4o to Logic Puzzle]
\small
\begin{Verbatim}[formatcom=\normalfont, breaklines=true, breaksymbolleft={}, breaksymbolright={}, commandchars=\\\{\}]
To solve the logic puzzle, let's analyze the clues and deduce the relationships between the variables.

### Step-by-Step Analysis:

1. **Clue 1**: The commuter who rides a 10-speed bike is not Bradley.
   - Bradley does not ride a 10-speed bike.

2. **Clue 2**: Kayla isn't related to Robert Duvall and doesn't ride a segway.
   - Kayla's actor is not Robert Duvall, and her conveyance is not a segway.

3. **Clue 3**: Robert Duvall's cousin has an anniversary after the commuter who rides a segway.
   - The segway rider's anniversary is before Robert Duvall's cousin.

4. **Clue 4**: Morgan Freeman's cousin is not Asher and doesn't ride a segway.
   - Morgan Freeman's cousin is not Asher and does not ride a segway.

5. **Clue 5**: Liam Neeson's cousin is Malia.
   - Malia's actor is Liam Neeson.

6. **Clue 6**: Liam Neeson's cousin uses a mountain bike to get to work.
   - Malia rides a mountain bike.

7. **Clue 7**: Of Malia and the commuter who rides a 10-speed bike, one has an anniversary on January 28 and the other has an anniversary on November 23.
   - Malia or the 10-speed bike rider has an anniversary on January 28 or November 23.

8. **Clue 8**: Either the person whose anniversary is on March 6 or the person whose anniversary is on November 2 is Morgan Freeman's cousin.
   - Morgan Freeman's cousin has an anniversary on March 6 or November 2.

9. **Clue 9**: Tom Cruise's cousin has an anniversary after the commuter who rides a skateboard.
   - The skateboard rider's anniversary is before Tom Cruise's cousin.

### Deductions:

- From Clue 5 and Clue 6, Malia rides a mountain bike and is Liam Neeson's cousin.

- From Clue 7, Malia's anniversary is either January 28 or November 23.

- From Clue 8, Morgan Freeman's cousin has an anniversary on March 6 or November 2.

- From Clue 9, the skateboard rider's anniversary is before Tom Cruise's cousin.

### Assignments:

- **Malia**: Since Malia rides a mountain bike and is Liam Neeson's cousin, she cannot be Morgan Freeman's cousin. Therefore, Malia's anniversary is not March 6 or November 2. Thus, Malia's anniversary is January 28 or November 23.

- **Kayla**: Kayla is not related to Robert Duvall and doesn't ride a segway. She could be Morgan Freeman's cousin with an anniversary on March 6 or November 2.

- **Bradley**: Bradley does not ride a 10-speed bike. He could be the skateboard rider.

- **Asher**: Asher is not Morgan Freeman's cousin, so he could be Tom Cruise's cousin.

### Final Solution:

Based on the deductions and assignments, the final solution is:

<Answer>
[
    \{
        "anniversaries": "January 28",
        "first names": "Malia",
        "conveyances": "mountain bike",
        "actors": "Liam Neeson"
    \},
    \{
        "anniversaries": "March 6",
        "first names": "Kayla",
        \textcolor{red}{"conveyances": "10-speed bike", \ding{55}}
        "actors": "Morgan Freeman"
    \},
    \{
        "anniversaries": "November 2",
        "first names": "Bradley",
        \textcolor{red}{"conveyances": "skateboard", \ding{55}}
        \textcolor{red}{"actors": "Robert Duvall \ding{55}}
    \},
    \{
        "anniversaries": "November 23",
        "first names": "Asher",
        \textcolor{red}{"conveyances": "segway", \ding{55}}
        \textcolor{red}{"actors": "Tom Cruise" \ding{55}}
    \}
]
</Answer>
\normalsize \textcolor{red}{\textit{\textbf{
- Only single deduction
- Insufficient evidence
- Lack of iterative deduction to ensure consistency with all constraints}}}
\end{Verbatim}
\end{tcolorbox}

\subsection{Stuck in Contradictions}
\label{stuck}
The following example shows that Llama-3.1-70B falls into repetition of the same sentences when discovering contradictions. Instead of backtracking and revising its assumptions when a contradiction is found (e.g., Cassidy cannot finish at both 3 min 2 sec and 3 min 8 sec), the model appears to get stuck in a loop, repeatedly stating the conflicting information without making progress.
\begin{tcolorbox}[breakable, colback=blue!8, title=Example 2: Response of Llama-3.1-70B to Logic Puzzle]
\small
\begin{Verbatim}[formatcom=\normalfont, breaklines=true, breaksymbolleft={}, breaksymbolright={}, commandchars=\\\{\}]
Let's start solving the logic puzzle step by step.

Participants and Attributes
Times: 2 min, 59 sec, 3 min, 2 sec, 3 min, 8 sec, 3 min, 14 sec
First Names: Cassidy, Eva, Grace, Savannah
Breads: challah, flatbread, focaccia, matzo
Fruits: currants, kiwis, oranges, pineapples

Applying Clue 1:
    Either the person who bought the matzo or Cassidy finished at 2 min, 59 sec and the other grows pineapples.
    Clue 8 states that Savannah finished at 2 min, 59 sec.
    Therefore, Cassidy grows pineapples and did not buy the matzo.

Applying Clue 8:
    Savannah finished at 2 min, 59 sec.
    Clue 6 states that Savannah finished before the person who bought the focaccia.
    Therefore, the person who bought the focaccia finished at 3 min, 2 sec, 3 min, 8 sec, or 3 min, 14 sec.

Applying Clue 4:
    The person who grows currants bought the focaccia.
    Clue 2 states that either the runner who finished at 3 min, 2 sec or 3 min, 8 sec grows currants.
    Therefore, the person who bought the focaccia finished at 3 min, 2 sec or 3 min, 8 sec.

Applying Clue 9:
    The person who grows currants finished before Grace.
    Since the person who bought the focaccia finished at 3 min, 2 sec or 3 min, 8 sec, Grace finished at 3 min, 14 sec.

Applying Clue 5:
    The runner who finished at 3 min, 14 sec did not buy the challah.
    Therefore, Grace bought the matzo.

Applying Clue 3:
    Cassidy did not finish at 3 min, 2 sec and did not buy the flatbread.
    Since Cassidy did not buy the matzo (Clue 1) and did not buy the flatbread, Cassidy bought the challah.

Applying Clue 7:
    The person who grows oranges is not Grace.
    Since Cassidy grows pineapples, the person who grows oranges is either Eva or Savannah.

Assigning Remaining Attributes:
    Eva or Savannah grows oranges.
    Since Savannah finished at 2 min, 59 sec and the person who bought the focaccia finished at 3 min, 2 sec or 3 min, 8 sec, Eva grows oranges and bought the flatbread.
    Savannah grows kiwis and bought the focaccia is incorrect because the person who bought the focaccia finished at 3 min, 2 sec or 3 min, 8 sec.
    Therefore, Savannah bought the challah is incorrect, and Eva bought the flatbread and grows oranges.
    The person who bought the focaccia finished at 3 min, 2 sec or 3 min, 8 sec.
    Since the person who grows currants finished before Grace, the person who bought the focaccia finished at 3 min, 2 sec.
    Therefore, the person who finished at 3 min, 8 sec grows pineapples is incorrect, and Cassidy finished at 3 min, 8 sec is incorrect.
    \textcolor{red}{Cassidy finished at 3 min, 14 sec is incorrect because Grace finished at 3 min, 14 sec.}
    \textcolor{red}{Therefore, Cassidy finished at 3 min, 2 sec is incorrect, and Cassidy finished at 3 min, 8 sec is incorrect.}
    \textcolor{red}{Cassidy finished at 3 min, 14 sec is incorrect because Grace finished at 3 min, 14 sec.}
    \textcolor{red}{Therefore, Cassidy finished at 3 min, 2 sec is incorrect, and Cassidy finished at 3 min, 8 sec is incorrect.}
    \textcolor{red}{... ...}
    \normalsize \textcolor{red}{\textit{\textbf{
    - Looping within the same sentence}}}

\end{Verbatim}
\end{tcolorbox}

The following example indicates that QwQ-32B struggles with contraction involving multiple interconnected variables. The primary issue with the reasoning is its disorganized and cyclical behavior. The model jumps between different deductions and assumptions without systematically eliminating possibilities or locking in firm conclusions.

\begin{tcolorbox}[breakable, colback=blue!8, title=Example 3: Response of QwQ-32B to Logic Puzzle]
\small
\begin{Verbatim}[formatcom=\normalfont, breaklines=true, breaksymbolleft={}, breaksymbolright={}, commandchars=\\\{\}]
Let's tackle this logic puzzle step by step. It's a bit complex with multiple variables, but I'll try to break it down systematically.

First, let's list out all the variables we have:

... ...

Now, let's look at the anniversaries.

... ...

\textcolor{red}{So, if Bradley has Mar 6}, Tom Cruise's cousin must have Nov 2.

\textcolor{red}{If Bradley has Nov 2}, Tom Cruise's cousin must have a later date, but Nov 23 is after Nov 2, but Nov 23 is already assigned to either Malia or Kayla.

Wait, anniversaries are Jan 28, Mar 6, Nov 2, Nov 23.

... ...

\textcolor{red}{So, if Bradley has Mar 6}, Tom Cruise's cousin has Nov 2.

\textcolor{red}{If Bradley has Nov 2}, Tom Cruise's cousin has Nov 23.

But Nov 23 is assigned to either Malia or Kayla.

... ...

So, \textcolor{red}{if Bradley has Nov 2}, Tom Cruise's cousin must have Nov 23, but Nov 23 is assigned to either Malia or Kayla.

Wait, that can't be, because Tom Cruise's cousin must be someone else.

... ...

\textcolor{red}{If Bradley has Mar 6}, then Tom Cruise's cousin has Nov 2.

... ...

\textcolor{red}{But if Bradley has Nov 2}, Tom Cruise's cousin must have Nov 23, which is assigned to Malia or Kayla.

This seems conflicting.

Wait, perhaps I need to consider the possible combinations.

Let's consider two scenarios:

\textcolor{red}{**Scenario 1: Bradley has Mar 6**}

... ...

Therefore, Scenario 1 is invalid.

\textcolor{red}{Let's consider **Scenario 2: Bradley has Nov 2**}

... ...

Therefore, this assignment is also invalid.

... ...

Given the complexity, I'll have to conclude with the best possible answer based on the steps I've taken.

... ...
\normalsize \textcolor{red}{\textit{\textbf{
- Stuck in a reasoning cycle
- Jumping between different deductions and assumptions}}}

\end{Verbatim}
\end{tcolorbox}

\subsection{Giving-up Moment}
\label{giveup}
The following examples show the "Giving-up Moment" in the responses of QwQ-32B and DeepSeek-R1 to the Cryptogram task. Both models, faced with the complexity of the problem and the constraints of inference time, reach a point where they give up further decryption attempts.

\begin{tcolorbox}[breakable, colback=blue!8, title=Example 4: Response of QwQ-32B to Cryptogram]
\small
\begin{Verbatim}[formatcom=\normalfont, breaklines=true, breaksymbolleft={}, breaksymbolright={}, commandchars=\\\{\}]
I'm going to try to solve this cryptogram step by step. The cryptogram is:

HNG UGEJ LROUKG RT EJXRLH EJJ ROU KUCXGL, CT HNG HUROAJG CL HEWGF HR HUEKG HNGX HR E KRXXRF RUCICF, YCJJ AG TROFQ HR AG CF CQJGFGLL.

— Walter Gaston Shotwell

First, I need to understand that this is a substitution cipher, where each letter in the original text is replaced by another letter. 

My goal is to figure out which letter corresponds to which in the original text.

... ...

But that might not be accurate.

Alternatively, "KRXXRF" could be "SELVES", so X = L, R = V, F = E.

But again, just a guess.

This is getting too speculative.

\textcolor{red}{Given the time constraints, I think I'll stop here and try to provide the best possible mapping and decryption based on my attempts.}

<Mapping>
\{"H": "T", "N": "H", "G": "E", "U": "O", "J": "A", "L": "R", "R": "B", "E": "I", "T": "S", "O": "C", "K": "D", "C": "M", "X": "P", "F": "U", "Q": "W", "A": "F", "W": "G", "S": "Y", "Y": "K", "I": "Z"\}
</Mapping>

<Answer>
THE OIA ARBORE BIS IAUPRH IAA COU MPORHE, CY THE THUOAFM CI HIWGF HB HUEKG THNG HB I KRXPPU MUCICF, YAJJ AF TROFQ HB AF CF CQJGFGLL. — Walter Gaston Shotwell
</Answer>
\normalsize \textcolor{red}{\textit{\textbf{
- Occurrence of "Giving-up Moment"}}}

\end{Verbatim}
\end{tcolorbox}

\begin{tcolorbox}[breakable, colback=blue!8, title=Example 5: Response of DeepSeek-R1 to Cryptogram]
\small
\begin{Verbatim}[formatcom=\normalfont, breaklines=true, breaksymbolleft={}, breaksymbolright={}, commandchars=\\\{\}]
<think>
Okay, let's tackle this cryptogram. The author is Charles Baudelaire, a French poet known for his symbolic and often darkly romantic style. That might not immediately help with common phrases, but maybe some themes like beauty, art, or suffering. Let's look at the cryptogram:

AM AR YIPYIMMSQWI MNSM, SLDBP MNI YAPNMR DV LSB, MNI YAPNM DV HDBMYSFAHMABP DBIRIWV NSR QIIB VDYPDMMIB. — Charles Baudelaire

First, I'll note the structure. The first part ends with a comma, then another phrase, and a longer sentence. Let's break it down into parts.

... ...

But "corruption" is 10 letters, YIPYIMMSQWI is 11 letters. So maybe with a comma: "Art is the corruption, the soul..."

But this is not helpful.

\textcolor{red}{Given the time I've spent and the need to provide an answer, I'll proceed to hypothesize the final answer based on common Baudelaire themes and possible mappings.}
</think>

<Mapping>
{'A': 'T', 'M': 'H', 'R': 'E', 'Y': 'B', 'I': 'U', 'P': 'A', 'S': 'L', 'Q': 'I', 'W': 'G', 'N': 'S', 'D': 'O', 'B': 'R', 'V': 'M', 'L': 'F', 'F': 'Y', 'H': 'P', 'G': 'C', 'C': 'D', 'K': 'K', 'X': 'X', 'Z': 'Z'}
</Mapping>

<Answer>
The beautiful is the shadow of the true.
— Charles Baudelaire
</Answer>
\normalsize \textcolor{red}{\textit{\textbf{
- Occurrence of "Giving-up Moment"}}}

\end{Verbatim}
\end{tcolorbox}

\begin{table*}[!h]
    \centering
    {\resizebox{0.8\textwidth}{!}{\begin{tabular}{lcrl}
        \toprule
        \textbf{Model} & \textbf{Architecture} & \textbf{Size} & \textbf{Huggingface/API} \\
        \midrule
        \rowcolor{gray!15} \multicolumn{4}{l}{\textit{\textbf{Open-source LLMs}}}\\
        \midrule 
        Llama-3.1-8B-Instruct \citep{dubey2024llama} & Dense & 8B & meta-llama/Llama-3.1-8B-Instruct \\
        Llama-3.1-70B-Instruct \citep{dubey2024llama} & Dense & 70B & meta-llama/Llama-3.1-70B-Instruct \\
        Llama-3.3-70B-Instruct \citep{dubey2024llama} & Dense & 70B & meta-llama/Llama-3.3-70B-Instruct \\
        Mistral-7B-Instruct-v0.3 \citep{jiang2023mistral} & Dense & 7B & mistralai/Mistral-7B-Instruct-v0.3 \\
        Mistral-Small-Instruct-2409 \citep{jiang2023mistral} & Dense & 22B & mistralai/Mistral-Small-Instruct-2409 \\
        Mistral-Large-Instruct-2411 \citep{jiang2023mistral} & Dense & 123B & mistralai/Mistral-Large-Instruct-2411 \\
        Qwen2.5-7B-Instruct \citep{yang2024qwen2} & Dense & 7B & Qwen/Qwen2.5-7B-Instruct \\
        Qwen2.5-32B-Instruct \citep{yang2024qwen2} & Dense & 32B & Qwen/Qwen2.5-32B-Instruct \\
        Qwen2.5-72B-Instruct \citep{yang2024qwen2} & Dense & 72B & Qwen/Qwen2.5-72B-Instruct \\
        QwQ-32B-Preview \citep{qwq-32b-preview} & Dense & 32B & Qwen/QwQ-32B-Preview \\
        DeepSeek-R1 \citep{guo2025deepseek} & MoE & 671B & deepseek-ai/DeepSeek-R1 \\
        \midrule
        \rowcolor{blue!15} \multicolumn{4}{l}{\textit{\textbf{Closed-source LLMs}}} \\
        Gemini-2.0-flash \citep{deepmind_gemini_flash} & - & - & \texttt{gemini-2.0-flash-exp} \\
        Gemini-2.0-flash-thinking \citep{deepmind_gemini_flash_thinking} & - & - & \texttt{gemini-2.0-flash-thinking-exp-1219} \\
        GPT-4o \citep{gpt4o} & - & - & \texttt{gpt-4o-2024-08-06} \\
        o1-mini \citep{o1} & - & - & \texttt{o1-mini-2024-09-12} \\
        o1-preview \citep{o1} & - & - & \texttt{o1-preview-2024-09-12} \\
        \bottomrule
    \end{tabular}}}
    \caption{\label{models}Detailed information of all evaluated models in LR${}^{2}$Bench.} 
\end{table*}
\begin{table*}[t]
    \centering
    \resizebox{0.9\textwidth}{!}{
    \setlength{\tabcolsep}{3pt}
    \begin{tabular}{l|ccccc|ccccc|ccccc}
        \toprule
        \multirow{2}{*}{\textbf{Model}} &  \multicolumn{5}{c|}{\textbf{Crossword - $5\times5$}} & \multicolumn{5}{c|}{\textbf{Crossword - $10\times10$}}  & \multicolumn{5}{c}{\textbf{Crossword - $15\times15$}} \\
        \cmidrule(l){2-16}
        & CR & S-Acc & EM & PM-0.5 & \# Tokens & CR & S-Acc & EM & PM-0.5 & \# Tokens & CR & S-Acc & EM & PM-0.5 & \# Tokens \\
        \midrule
        \rowcolor{gray!15} \multicolumn{16}{l}{\textit{\textbf{Open-source LLMs}}}\\
        \midrule 
Llama-3.1-8B-Instruct & 64.0 & 29.6 & 0.0 & 32.0 & 743 & 50.0 & 19.2 & 0.0 & 10.0 & 2433 & 70.0 & 21.2 & 0.0 & 0.0 & 5488 \\
Llama-3.1-70B-Instruct & 72.0 & 46.6 & 0.0 & 64.0 & 724 & 74.0 & 47.8 & 0.0 & 66.0 & 2554 & 86.0 & 46.1 & 0.0 & 56.0 & 5938 \\
Llama-3.3-70B-Instruct & {\ul 92.0} & 59.2 & 0.0 & 78.0 & 817 & {\ul 78.0} & 48.5 & 0.0 & 72.0 & 2284 & 86.0 & 35.3 & 0.0 & 46.0 & 4738 \\
Mistral-7B-Instruct-v0.3 & \textbf{100.0} & 27.2 & 0.0 & 18.0 & 970 & \textbf{100.0} & 27.9 & 0.0 & 2.0 & 3134 & 82.0 & 13.8 & 0.0 & 0.0 & 6863 \\
Mistral-Small-Instruct-2409 & \textbf{100.0} & 52.8 & 0.0 & 66.0 & 803 & \textbf{100.0} & 52.0 & 0.0 & 64.0 & 2469 & 96.0 & 40.0 & 0.0 & 32.0 & 6133 \\
Mistral-Large-Instruct-2411 & \textbf{100.0} & {\ul 68.8} & {\ul 6.0} & {\ul 92.0} & 804 & \textbf{100.0} & {\ul 66.2} & 0.0 & 96.0 & 2634 & {\ul 98.0} & {\ul 53.2} & 0.0 & {\ul 70.0} & 6273 \\
Qwen2.5-7B-Instruct & \textbf{100.0} & 27.8 & 0.0 & 10.0 & 659 & \textbf{100.0} & 21.7 & 0.0 & 0.0 & 2022 & 96.0 & 13.8 & 0.0 & 0.0 & 4643 \\
Qwen2.5-32B-Instruct & \textbf{100.0} & 42.6 & 0.0 & 44.0 & 678 & \textbf{100.0} & 37.8 & 0.0 & 16.0 & 2132 & \textbf{100.0} & 23.4 & 0.0 & 0.0 & 4871 \\
Qwen2.5-72B-Instruct & \textbf{100.0} & 51.0 & 0.0 & 64.0 & 720 & \textbf{100.0} & 47.0 & 0.0 & 44.0 & 2167 & \textbf{100.0} & 34.4 & 0.0 & 2.0 & 5318 \\
QwQ-32B-Preview & 80.0 & 35.2 & 0.0 & 32.0 & 2965 & {\ul 78.0} & 31.9 & 0.0 & 22.0 & 4614 & 82.0 & 23.6 & 0.0 & 0.0 & 6872 \\
DeepSeek-R1 & \textbf{100.0} & \textbf{87.2} & \textbf{50.0} & \textbf{98.0} & 8565 & \textbf{100.0} & \textbf{78.5} & 0.0 & \textbf{100.0} & 10314 & \textbf{100.0} & \textbf{60.4} & 0.0 & \textbf{84.0} & 10552 \\
        \midrule
        \rowcolor{blue!15} \multicolumn{16}{l}{\textit{\textbf{Closed-source LLMs}}}\\
        \midrule
Gemini-2.0-flash & \textbf{100.0} & 65.6 & 0.0 & 84.0 & 622 & \textbf{100.0} & 69.4 & 0.0 & {\ul 98.0} & 2079 & {\ul 96.0} & {\ul 49.8} & 0.0 & {\ul 68.0} & 4964 \\
Gemini-2.0-flash-thinking & {\ul 98.0} & 65.0 & {\ul 4.0} & 90.0 & 744 & 98.0 & 64.8 & 0.0 & 92.0 & 2241 & 88.0 & 43.4 & 0.0 & 56.0 & 4961 \\
GPT-4o & \textbf{100.0} & {\ul 73.2} & {\ul 4.0} & {\ul 96.0} & 597 & \textbf{100.0} & {\ul 74.1} & 0.0 & {\ul 98.0} & 1679 & \textbf{100.0} & 41.7 & 0.0 & 66.0 & 2902 \\
o1-mini & \textbf{100.0} & 58.8 & {\ul 4.0} & 80.0 & 7243 & 92.0 & 49.0 & 0.0 & 68.0 & 7292 & 94.0 & 28.6 & 0.0 & 14.0 & 8986 \\
o1-preview & {\ul 98.0} & \textbf{95.1} & \textbf{74.0} & \textbf{98.0} & 8006 & \textbf{100.0} & \textbf{80.8} & 0.0 & \textbf{100.0} & 9971 & {\ul 96.0} & \textbf{57.1} & 0.0 & \textbf{70.0} & 12317 \\
        \bottomrule
    \end{tabular}}    
    \caption{Performance (\%) of LLMs on Crossword across all difficulty levels. The best and second-best results are highlighted in \textbf{bold} and \underline{underlined}, respectively. "\# Tokens" denotes the average number of generated tokens.}
    \label{level_crossword}
\end{table*}
\begin{table*}[t]
    \centering
    \resizebox{1.0\textwidth}{!}{
    \setlength{\tabcolsep}{3pt}
    \begin{tabular}{l|ccccc|ccccc|ccccc|ccccc}
        \toprule
        \multirow{2}{*}{\textbf{Model}} &  \multicolumn{5}{c|}{\textbf{Acrostic - Easy}} & \multicolumn{5}{c|}{\textbf{Acrostic - Hard}} & \multicolumn{5}{c|}{\textbf{Cryptogram - Easy}} & \multicolumn{5}{c|}{\textbf{Cryptogram - Hard}}\\
        \cmidrule(l){2-21}
        & CR & S-Acc & EM & PM-0.5 & \# Tokens & CR & S-Acc & EM & PM-0.5 & \# Tokens & CR & S-Acc & EM & PM-0.5 & \# Tokens & CR & S-Acc & EM & PM-0.5 & \# Tokens \\
        \midrule
        \rowcolor{gray!15} \multicolumn{21}{l}{\textit{\textbf{Open-source LLMs}}}\\
        \midrule 
Llama-3.1-8B-Instruct & 56.0 & 8.9 & 0.0 & 0.0 & 2587 & 30.0 & 2.2 & 0.0 & 0.0 & 4838 & 42.0 & 2.3 & 0.0 & 0.0 & 1977 & 44.0 & 2.2 & 0.0 & 0.0 & 2159 \\
Llama-3.1-70B-Instruct & 88.0 & 41.6 & 0.0 & 34.0 & 2504 & 80.0 & 30.0 & 0.0 & 8.0 & 4626 & 64.0 & 7.7 & 0.0 & 0.0 & 1465 & 60.0 & 6.0 & 0.0 & {\ul 2.0} & 1130 \\
Llama-3.3-70B-Instruct & \textbf{100.0} & {\ul 46.6} & 0.0 & {\ul 46.0} & 2514 & {\ul 94.0} & 35.0 & 0.0 & {\ul 10.0} & 4655 & {\ul 98.0} & 16.3 & 0.0 & 0.0 & 1135 & \textbf{100.0} & {\ul 12.3} & 0.0 & {\ul 2.0} & 1140 \\
Mistral-7B-Instruct-v0.3 & 84.0 & 9.5 & 0.0 & 0.0 & 3097 & 66.0 & 6.2 & 0.0 & 0.0 & 6103 & {\ul 98.0} & 3.4 & 0.0 & 0.0 & 1179 & \textbf{100.0} & 5.1 & 0.0 & 0.0 & 1012 \\
Mistral-Small-Instruct-2409 & 84.0 & 8.1 & 0.0 & 0.0 & 2839 & 50.0 & 2.9 & 0.0 & 0.0 & 5503 & 94.0 & 8.4 & 0.0 & 0.0 & 1307 & 96.0 & 5.6 & 0.0 & 0.0 & 1159 \\
Mistral-Large-Instruct-2411 & {\ul 96.0} & 44.8 & 0.0 & 36.0 & 2892 & \textbf{100.0} & 34.0 & 0.0 & 4.0 & 5667 & 94.0 & {\ul 18.3} & 0.0 & {\ul 2.0} & 1250 & {\ul 98.0} & 9.1 & 0.0 & 0.0 & 1158 \\
Qwen2.5-7B-Instruct & 58.0 & 5.5 & 0.0 & 0.0 & 2896 & 26.0 & 1.7 & 0.0 & 0.0 & 5422 & 84.0 & 4.8 & 0.0 & 0.0 & 1205 & 78.0 & 2.1 & 0.0 & 0.0 & 1158 \\
Qwen2.5-32B-Instruct & \textbf{100.0} & 32.3 & 0.0 & 2.0 & 2763 & \textbf{100.0} & 31.2 & 0.0 & 2.0 & 5383 & 88.0 & 9.9 & 0.0 & 0.0 & 1404 & 90.0 & 9.8 & 0.0 & 0.0 & 1202 \\
Qwen2.5-72B-Instruct & \textbf{100.0} & 41.7 & 0.0 & 30.0 & 2793 & \textbf{100.0} & {\ul 36.9} & 0.0 & 6.0 & 5429 & 86.0 & 14.9 & 0.0 & 0.0 & 1743 & 84.0 & 8.6 & 0.0 & 0.0 & 1710 \\
QwQ-32B-Preview & \textbf{100.0} & 36.4 & 0.0 & 12.0 & 3380 & {\ul 94.0} & 26.9 & 0.0 & 0.0 & 6549 & 42.0 & 4.1 & 0.0 & 0.0 & 6432 & 52.0 & 3.1 & 0.0 & 0.0 & 6551 \\
DeepSeek-R1 & \textbf{100.0} & \textbf{66.5} & 0.0 & \textbf{90.0} & 9095 & \textbf{100.0} & \textbf{57.8} & 0.0 & \textbf{76.0} & 11058 & \textbf{100.0} & \textbf{36.7} & \textbf{6.0} & \textbf{32.0} & 10404 & \textbf{100.0} & \textbf{15.3} & \textbf{2.0} & \textbf{10.0} & 10284 \\
        \midrule
        \rowcolor{blue!15} \multicolumn{21}{l}{\textit{\textbf{Closed-source LLMs}}}\\
        \midrule
Gemini-2.0-flash & \textbf{100.0} & 50.6 & 0.0 & 60.0 & 2722 & {\ul 96.0} & 45.4 & 0.0 & 36.0 & 5318 & 42.0 & 10.3 & 0.0 & 2.0 & 1844 & 52.0 & 6.7 & 0.0 & 0.0 & 1327 \\
Gemini-2.0-flash-thinking & \textbf{100.0} & 46.2 & 0.0 & 36.0 & 3027 & 84.0 & 35.3 & 0.0 & 18.0 & 5486 & 68.0 & 14.9 & 0.0 & 4.0 & 4244 & 68.0 & 7.5 & 0.0 & 0.0 & 4090 \\
GPT-4o & \textbf{100.0} & {\ul 58.5} & 0.0 & {\ul 74.0} & 2288 & \textbf{100.0} & {\ul 53.5} & 0.0 & {\ul 60.0} & 4171 & \textbf{100.0} & 25.4 & 0.0 & 8.0 & 750 & \textbf{100.0} & {\ul 15.9} & 0.0 & 2.0 & 729 \\
o1-mini & 98.0 & 37.5 & 0.0 & 24.0 & 10018 & {\ul 96.0} & 31.8 & 0.0 & 0.0 & 11885 & \textbf{100.0} & {\ul 30.9} & {\ul 2.0} & {\ul 20.0} & 11689 & \textbf{100.0} & 14.5 & 0.0 & {\ul 6.0} & 10727 \\
o1-preview & \textbf{100.0} & \textbf{68.3} & 0.0 & \textbf{90.0} & 13096 & \textbf{100.0} & \textbf{66.2} & 0.0 & \textbf{90.0} & 16598 & {\ul 96.0} & \textbf{48.9} & \textbf{18.0} & \textbf{42.0} & 12096 & {\ul 88.0} & \textbf{20.7} & \textbf{8.0} & \textbf{16.0} & 13039 \\
        \bottomrule
    \end{tabular}}    
    \caption{Performance (\%) of LLMs on Acrostic and Cryptogram across all difficulty levels. The best and second-best results are highlighted in \textbf{bold} and \underline{underlined}, respectively. "\# Tokens" denotes the average number of generated tokens.}
    \label{level_acrostic_cryptogram}
\end{table*}
\begin{table*}[t]
    \centering
    \resizebox{1.0\textwidth}{!}{
    \setlength{\tabcolsep}{3pt}
    \begin{tabular}{l|ccccc|ccccc|ccccc|ccccc}
        \toprule
        \multirow{2}{*}{\textbf{Model}} &  \multicolumn{5}{c|}{\textbf{Logic Puzzle - $4\times4$}} & \multicolumn{5}{c|}{\textbf{Logic Puzzle - $4\times5$}}  & \multicolumn{5}{c}{\textbf{Logic Puzzle - $4\times6$}} & \multicolumn{5}{c}{\textbf{Logic Puzzle - $4\times7$}} \\
        \cmidrule(l){2-21}
        & CR & S-Acc & EM & PM-0.5 & \# Tokens & CR & S-Acc & EM & PM-0.5 & \# Tokens & CR & S-Acc & EM & PM-0.5 & \# Tokens & CR & S-Acc & EM & PM-0.5 & \# Tokens \\
        \midrule
        \rowcolor{gray!15} \multicolumn{21}{l}{\textit{\textbf{Open-source LLMs}}}\\
        \midrule 
Llama-3.1-8B-Instruct & 86.0 & 30.7 & 0.0 & 28.0 & 1049 & 52.0 & 12.8 & 0.0 & 0.0 & 1214 & 46.0 & 10.6 & 0.0 & 2.0 & 1225 & 44.0 & 9.8 & 0.0 & 2.0 & 1683 \\
Llama-3.1-70B-Instruct & 66.0 & 35.2 & 8.0 & 42.0 & 989 & 58.0 & 20.0 & 0.0 & 10.0 & 1126 & 54.0 & 18.9 & 0.0 & 12.0 & 1109 & 46.0 & 17.1 & 0.0 & 8.0 & 1435 \\
Llama-3.3-70B-Instruct & 88.0 & 46.7 & 4.0 & 52.0 & 1258 & 82.0 & 35.6 & 0.0 & 26.0 & 1576 & 90.0 & 25.4 & 0.0 & 12.0 & 1931 & 62.0 & 21.2 & 0.0 & 10.0 & 2188 \\
Mistral-7B-Instruct-v0.3 & \textbf{100.0} & 26.8 & 0.0 & 16.0 & 1289 & {\ul 98.0} & 18.0 & 0.0 & 2.0 & 1590 & 96.0 & 15.0 & 0.0 & 0.0 & 1846 & {\ul 94.0} & 16.8 & 0.0 & 0.0 & 1746 \\
Mistral-Small-Instruct-2409 & \textbf{100.0} & 38.7 & 2.0 & 34.0 & 1358 & \textbf{100.0} & 33.6 & 0.0 & 12.0 & 1424 & {\ul 98.0} & 23.7 & 0.0 & 2.0 & 1554 & \textbf{100.0} & 26.7 & 0.0 & 2.0 & 1721 \\
Mistral-Large-Instruct-2411 & \textbf{100.0} & 53.2 & 10.0 & 62.0 & 1293 & \textbf{100.0} & 40.4 & 2.0 & 30.0 & 1532 & \textbf{100.0} & {\ul 30.6} & 0.0 & 18.0 & 1772 & \textbf{100.0} & {\ul 29.1} & 0.0 & 12.0 & 1950 \\
Qwen2.5-7B-Instruct & \textbf{100.0} & 33.5 & 0.0 & 20.0 & 1133 & {\ul 98.0} & 28.8 & 0.0 & 14.0 & 1254 & 96.0 & 21.0 & 0.0 & 0.0 & 1494 & 92.0 & 19.7 & 0.0 & 0.0 & 1701 \\
Qwen2.5-32B-Instruct & \textbf{100.0} & 45.5 & 0.0 & 54.0 & 1070 & 92.0 & 34.3 & 0.0 & 20.0 & 1137 & 94.0 & 24.1 & 0.0 & 6.0 & 1241 & 86.0 & 25.0 & 0.0 & 10.0 & 1383 \\
Qwen2.5-72B-Instruct & {\ul 94.0} & 48.3 & 0.0 & 56.0 & 1490 & {\ul 98.0} & 34.8 & 0.0 & 16.0 & 1641 & 92.0 & 26.3 & 0.0 & 14.0 & 1931 & 90.0 & 26.7 & 0.0 & 6.0 & 2177 \\
QwQ-32B-Preview & {\ul 94.0} & {\ul 76.3} & {\ul 56.0} & {\ul 80.0} & 4766 & 86.0 & {\ul 52.7} & {\ul 14.0} & {\ul 64.0} & 8966 & 68.0 & 30.2 & {\ul 4.0} & {\ul 28.0} & 11292 & 66.0 & 25.8 & {\ul 4.0} & {\ul 20.0} & 13070 \\
DeepSeek-R1 & \textbf{100.0} & \textbf{94.2} & \textbf{90.0} & \textbf{94.0} & 4724 & \textbf{100.0} & \textbf{70.5} & \textbf{36.0} & \textbf{68.0} & 8907 & \textbf{100.0} & \textbf{50.2} & \textbf{16.0} & \textbf{50.0} & 11860 & \textbf{100.0} & \textbf{62.6} & \textbf{28.0} & \textbf{60.0} & 11329 \\
        \midrule
        \rowcolor{blue!15} \multicolumn{21}{l}{\textit{\textbf{Closed-source LLMs}}}\\
        \midrule
Gemini-2.0-flash & 70.0 & 35.3 & 6.0 & 36.0 & 1569 & 56.0 & 22.8 & 2.0 & 16.0 & 2283 & 62.0 & 23.8 & 0.0 & 18.0 & 2254 & 44.0 & 14.8 & 0.0 & 10.0 & 2308 \\
Gemini-2.0-flash-thinking & \textbf{100.0} & 63.8 & 28.0 & 66.0 & 3310 & {\ul 98.0} & 44.0 & 2.0 & 34.0 & 4073 & \textbf{100.0} & 37.8 & 0.0 & 30.0 & 4311 & {\ul 98.0} & 38.1 & 2.0 & 20.0 & 4458 \\
GPT-4o & \textbf{100.0} & 55.0 & 14.0 & 62.0 & 871 & \textbf{100.0} & 41.9 & 0.0 & 30.0 & 907 & \textbf{100.0} & 29.9 & 0.0 & 14.0 & 974 & \textbf{100.0} & 30.3 & 0.0 & 12.0 & 1061 \\
o1-mini & {\ul 98.0} & {\ul 79.5} & {\ul 66.0} & {\ul 82.0} & 5572 & \textbf{100.0} & {\ul 60.5} & {\ul 16.0} & {\ul 62.0} & 9522 & {\ul 98.0} & {\ul 42.9} & {\ul 4.0} & {\ul 36.0} & 12121 & \textbf{100.0} & {\ul 45.9} & {\ul 8.0} & {\ul 34.0} & 13753 \\
o1-preview & \textbf{100.0} & \textbf{96.7} & \textbf{92.0} & \textbf{98.0} & 6199 & {\ul 98.0} & \textbf{74.5} & \textbf{48.0} & \textbf{74.0} & 9129 & \textbf{100.0} & \textbf{53.8} & \textbf{14.0} & \textbf{56.0} & 11140 & {\ul 98.0} & \textbf{50.1} & \textbf{10.0} & \textbf{46.0} & 11330 \\
        \bottomrule
    \end{tabular}}    
    \caption{Performance (\%) of LLMs on Logic Puzzle across all difficulty levels. The best and second-best results are highlighted in \textbf{bold} and \underline{underlined}, respectively. "\# Tokens" denotes the average number of generated tokens.}
    \label{level_logic}
\end{table*}
\begin{table*}[t]
    \centering
    \resizebox{1.0\textwidth}{!}{
    \setlength{\tabcolsep}{3pt}
    \begin{tabular}{l|ccccc|ccccc|ccccc|ccccc}
        \toprule
        \multirow{2}{*}{\textbf{Model}} &  \multicolumn{5}{c|}{\textbf{Sudoku - $4\times4$, Easy}} & \multicolumn{5}{c|}{\textbf{Sudoku - $4\times4$, Hard}}  & \multicolumn{5}{c}{\textbf{Sudoku - $9\times9$, Easy}} & \multicolumn{5}{c}{\textbf{Sudoku - $9\times9$, Hard}} \\
        \cmidrule(l){2-21}
        & CR & S-Acc & EM & PM-0.5 & \# Tokens & CR & S-Acc & EM & PM-0.5 & \# Tokens & CR & S-Acc & EM & PM-0.5 & \# Tokens & CR & S-Acc & EM & PM-0.5 & \# Tokens \\
        \midrule
        \rowcolor{gray!15} \multicolumn{21}{l}{\textit{\textbf{Open-source LLMs}}}\\
        \midrule 
Llama-3.1-8B-Instruct & 4.0 & 0.8 & 0.0 & 0.0 & 931 & 6.0 & 1.3 & 0.0 & 0.0 & 1468 & 12.0 & 1.6 & 0.0 & 0.0 & 5051 & 8.0 & 1.2 & 0.0 & 0.0 & 3678 \\
Llama-3.1-70B-Instruct & 68.0 & 32.4 & 0.0 & 38.0 & 1142 & 60.0 & 31.5 & 4.0 & 32.0 & 1343 & 74.0 & 16.4 & 0.0 & 0.0 & 2514 & 76.0 & 16.7 & 0.0 & 0.0 & 2761 \\
Llama-3.3-70B-Instruct & 84.0 & 44.2 & 10.0 & 48.0 & 1253 & 90.0 & 48.9 & 18.0 & 42.0 & 1200 & \textbf{100.0} & 24.7 & 0.0 & 0.0 & 917 & \textbf{100.0} & 21.5 & 0.0 & 0.0 & 877 \\
Mistral-7B-Instruct-v0.3 & 96.0 & 19.6 & 0.0 & 4.0 & 1333 & {\ul 92.0} & 17.5 & 0.0 & 2.0 & 1413 & 78.0 & 6.0 & 0.0 & 0.0 & 4725 & 70.0 & 4.4 & 0.0 & 0.0 & 4962 \\
Mistral-Small-Instruct-2409 & \textbf{100.0} & 33.2 & 0.0 & 18.0 & 1351 & \textbf{100.0} & 30.2 & 2.0 & 12.0 & 1272 & 80.0 & 9.5 & 0.0 & 0.0 & 2400 & 76.0 & 9.1 & 0.0 & 0.0 & 2850 \\
Mistral-Large-Instruct-2411 & 94.0 & {\ul 68.2} & 26.0 & {\ul 78.0} & 1204 & {\ul 92.0} & 54.2 & 14.0 & 54.0 & 1235 & 86.0 & 21.0 & 0.0 & 2.0 & 2910 & 70.0 & 14.5 & 0.0 & 0.0 & 2472 \\
Qwen2.5-7B-Instruct & {\ul 98.0} & 42.0 & 2.0 & 42.0 & 937 & \textbf{100.0} & 36.2 & 4.0 & 16.0 & 951 & {\ul 96.0} & 25.5 & 0.0 & 2.0 & 2022 & 84.0 & 17.1 & 0.0 & 0.0 & 2034 \\
Qwen2.5-32B-Instruct & \textbf{100.0} & 53.8 & 12.0 & 60.0 & 1095 & \textbf{100.0} & 50.2 & 2.0 & 46.0 & 1158 & \textbf{100.0} & {\ul 34.9} & 0.0 & {\ul 8.0} & 1435 & \textbf{100.0} & {\ul 32.5} & 0.0 & {\ul 8.0} & 1119 \\
Qwen2.5-72B-Instruct & \textbf{100.0} & 60.8 & 16.0 & 72.0 & 1380 & \textbf{100.0} & 55.6 & 6.0 & 58.0 & 1520 & 94.0 & 28.4 & 0.0 & 2.0 & 2543 & {\ul 96.0} & 27.3 & 0.0 & 4.0 & 2610 \\
QwQ-32B-Preview & 66.0 & 63.8 & {\ul 60.0} & 64.0 & 3217 & 76.0 & {\ul 72.0} & {\ul 66.0} & {\ul 72.0} & 5119 & 44.0 & 15.3 & 0.0 & 6.0 & 12575 & 32.0 & 9.5 & 0.0 & 0.0 & 12613 \\
DeepSeek-R1 & \textbf{100.0} & \textbf{100.0} & \textbf{100.0} & \textbf{100.0} & 2878 & \textbf{100.0} & \textbf{100.0} & \textbf{100.0} & \textbf{100.0} & 3620 & \textbf{100.0} & \textbf{46.6} & 0.0 & \textbf{42.0} & 14491 & \textbf{100.0} & \textbf{34.5} & 0.0 & \textbf{14.0} & 12117 \\
        \midrule
        \rowcolor{blue!15} \multicolumn{21}{l}{\textit{\textbf{Closed-source LLMs}}}\\
        \midrule
Gemini-2.0-flash & {\ul 94.0} & 57.8 & 24.0 & 66.0 & 1437 & 92.0 & 58.9 & 26.0 & 60.0 & 1454 & {\ul 94.0} & 33.8 & 0.0 & 10.0 & 4194 & {\ul 92.0} & 30.6 & 0.0 & {\ul 14.0} & 4287 \\
Gemini-2.0-flash-thinking & 70.0 & 62.8 & 44.0 & 68.0 & 2009 & 66.0 & 50.0 & 22.0 & 54.0 & 2350 & 92.0 & \textbf{41.7} & 0.0 & \textbf{36.0} & 5511 & 90.0 & {\ul 31.3} & 0.0 & 6.0 & 5541 \\
GPT-4o & \textbf{100.0} & 68.0 & 26.0 & 84.0 & 1015 & \textbf{100.0} & 63.3 & 32.0 & 64.0 & 1039 & \textbf{100.0} & {\ul 39.7} & 0.0 & {\ul 26.0} & 1234 & \textbf{100.0} & \textbf{37.9} & 0.0 & \textbf{18.0} & 1127 \\
o1-mini & \textbf{100.0} & {\ul 85.4} & {\ul 66.0} & {\ul 88.0} & 3059 & {\ul 96.0} & {\ul 68.6} & {\ul 42.0} & {\ul 72.0} & 3403 & \textbf{100.0} & 32.9 & 0.0 & 10.0 & 4428 & \textbf{100.0} & 26.7 & 0.0 & 2.0 & 4953 \\
o1-preview & \textbf{100.0} & \textbf{100.0} & \textbf{100.0} & \textbf{100.0} & 5616 & \textbf{100.0} & \textbf{100.0} & \textbf{100.0} & \textbf{100.0} & 6191 & 86.0 & 34.5 & 0.0 & 16.0 & 10328 & 80.0 & 25.7 & 0.0 & 6.0 & 10111 \\
        \bottomrule
    \end{tabular}}    
    \caption{Performance (\%) of LLMs on Sudoku across all difficulty levels. The best and second-best results are highlighted in \textbf{bold} and \underline{underlined}, respectively. "\# Tokens" denotes the average number of generated tokens.}
    \label{level_sudoku}
\end{table*}
\begin{table*}[t]
    \centering
    \resizebox{0.7\textwidth}{!}{
    \setlength{\tabcolsep}{3pt}
    \begin{tabular}{l|ccccc|ccccc}
        \toprule
        \multirow{2}{*}{\textbf{Model}} &  \multicolumn{5}{c|}{\textbf{Drop Quote - Easy}} & \multicolumn{5}{c|}{\textbf{Drop Quote - Hard}} \\
        \cmidrule(l){2-11}
        & CR & S-Acc & EM & PM-0.5 & \# Tokens & CR & S-Acc & EM & PM-0.5 & \# Tokens \\
        \midrule
        \rowcolor{gray!15} \multicolumn{11}{l}{\textit{\textbf{Open-source LLMs}}}\\
        \midrule 
Llama-3.1-8B-Instruct & 38.0 & 9.6 & 0.0 & 2.0 & 1692 & 50.0 & 12.8 & 0.0 & 0.0 & 2553 \\
Llama-3.1-70B-Instruct & 86.0 & 32.7 & 0.0 & 22.0 & 1298 & 78.0 & 22.7 & 0.0 & {\ul 2.0} & 1698 \\
Llama-3.3-70B-Instruct & {\ul 98.0} & 33.1 & 0.0 & 24.0 & 900 & \textbf{100.0} & 24.9 & 0.0 & {\ul 2.0} & 936 \\
Mistral-7B-Instruct-v0.3 & 72.0 & 8.9 & 0.0 & 2.0 & 2193 & 60.0 & 4.3 & 0.0 & 0.0 & 2480 \\
Mistral-Small-Instruct-2409 & {\ul 98.0} & 30.5 & 0.0 & 12.0 & 1357 & 96.0 & 23.3 & 0.0 & 0.0 & 1873 \\
Mistral-Large-Instruct-2411 & {\ul 98.0} & 29.4 & 0.0 & 18.0 & 1429 & {\ul 98.0} & 20.0 & 0.0 & 0.0 & 1702 \\
Qwen2.5-7B-Instruct & \textbf{100.0} & 24.9 & 0.0 & 8.0 & 1500 & 96.0 & 18.9 & 0.0 & 0.0 & 2204 \\
Qwen2.5-32B-Instruct & {\ul 98.0} & 33.5 & 0.0 & {\ul 26.0} & 1084 & 92.0 & 23.3 & 0.0 & {\ul 2.0} & 1310 \\
Qwen2.5-72B-Instruct & 96.0 & \textbf{35.5} & 0.0 & 24.0 & 1505 & 92.0 & {\ul 26.2} & 0.0 & {\ul 2.0} & 2009 \\
QwQ-32B-Preview & 32.0 & 9.6 & 0.0 & 14.0 & 5987 & 34.0 & 5.4 & 0.0 & {\ul 2.0} & 6169 \\
DeepSeek-R1 & \textbf{100.0} & \textbf{54.6} & \textbf{14.0} & \textbf{58.0} & 11202 & \textbf{100.0} & \textbf{40.1} & 0.0 & \textbf{26.0} & 11643 \\
        \midrule
        \rowcolor{blue!15} \multicolumn{11}{l}{\textit{\textbf{Closed-source LLMs}}}\\
        \midrule
Gemini-2.0-flash & 92.0 & 37.2 & 0.0 & 28.0 & 2149 & 92.0 & \textbf{31.3} & 0.0 & 6.0 & 3286 \\
Gemini-2.0-flash-thinking & {\ul 96.0} & 38.8 & 0.0 & {\ul 38.0} & 3621 & {\ul 96.0} & {\ul 30.0} & 0.0 & {\ul 8.0} & 3150 \\
GPT-4o & \textbf{98.0} & 34.8 & 0.0 & 24.0 & 1125 & \textbf{100.0} & 27.3 & 0.0 & 4.0 & 1205 \\
o1-mini & \textbf{98.0} & {\ul 41.4} & {\ul 4.0} & 36.0 & 14130 & 94.0 & 27.1 & 0.0 & 6.0 & 12380 \\
o1-preview & \textbf{98.0} & \textbf{49.6} & \textbf{18.0} & \textbf{56.0} & 13064 & {\ul 96.0} & 27.9 & \textbf{8.0} & \textbf{20.0} & 14126 \\
        \bottomrule
    \end{tabular}}    
    \caption{Performance (\%) of LLMs on Drop Quote across all difficulty levels. The best and second-best results are highlighted in \textbf{bold} and \underline{underlined}, respectively. "\# Tokens" denotes the average number of generated tokens.}
    \label{level_drop}
\end{table*}

\end{document}